\documentclass[11pt]{article}
\usepackage{archon}
\usepackage{placeins} %
\usepackage{needspace} %
\usepackage{pifont} %
\newlength{\orglogoheight}
\title{
Anchor 1:
Precise and Stable Humanoid Control for Whole-Body Manipulation
}
\orglabel{}
\orglogo{%
  \makebox[\linewidth][l]{%
    \includegraphics[height=0.8cm,keepaspectratio]{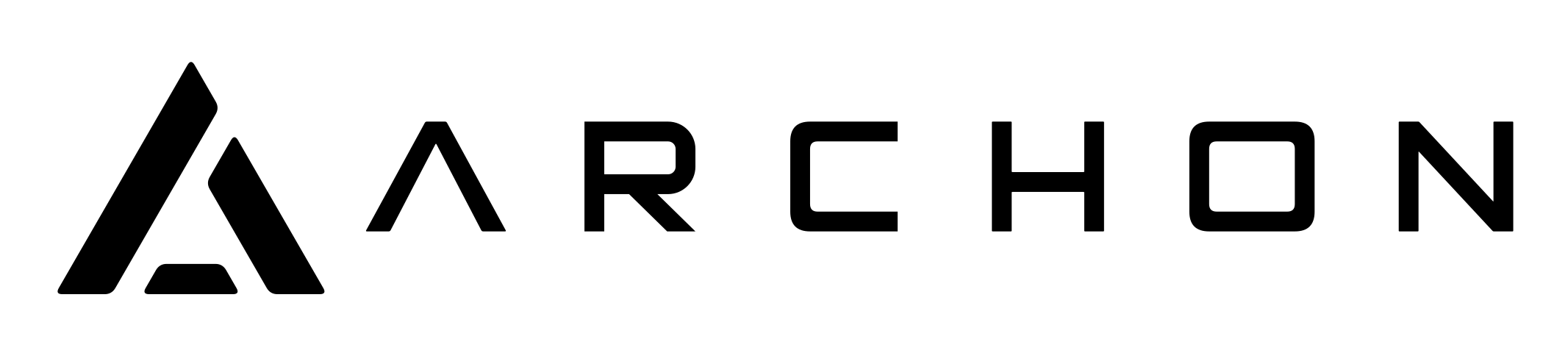}%
  }%
}

\contributors{%
  Archon Robotics\textsuperscript{*}
} 

\contriblegend{
    \textsuperscript{*}\,For detailed authorship, see Contributions section. 
}

\projectlinks{
{%
    \url{https://archon.tech/research/anchor1}%
}%
}

\newcommand{\bC}{\mathbf{C}}
\newcommand{\bd}{\mathbf{d}}

\newcommand{\bp}{\mathbf{p}}
\newcommand{\bq}{\mathbf{q}}
\newcommand{\bR}{\mathbf{R}}

\crefname{section}{Sec.}{Secs.}
\Crefname{section}{Section}{Sections}
\Crefname{figure}{Figure}{Figures}
\crefname{figure}{Fig.}{Figs.}
\Crefname{table}{Table}{Tables}
\crefname{table}{Table}{Tables}
\crefname{equation}{Eq.}{Eq.}
\Crefname{equation}{Equation}{Equations}

\makeatletter
\DeclareRobustCommand\onedot{\futurelet\@let@token\@onedot}
\def\@onedot{\ifx\@let@token.\else.\null\fi\xspace}

\makeatother

\definecolor{darkgreen}{rgb}{0,0.7,0}
\definecolor{darkblue}{RGB}{31,119,180}
\definecolor{darkred}{RGB}{214,39,40}
\definecolor{mediumgray}{rgb}{0.5,0.5,0.5}
\definecolor{mediumteal}{rgb}{0,0.5,0.5}

\definecolor{ellisred}{rgb}{0.87,0.44,0.38} %
\definecolor{ellisgreen}{rgb}{0.69,0.90,0.52} %
\definecolor{elliscyan}{rgb}{0.29,0.77,0.74} %
\definecolor{ellisorange}{rgb}{0.89,0.55,0.28} %
\definecolor{ellisblue}{rgb}{0.41,0.61,0.86} %

\newcommand{\cmark}{\ding{51}}
\newcommand{\xmark}{\ding{55}}

\newcommand{\modelname}{Anchor 1\xspace}

\begin{document}
\maketitle

\begin{abstract}
\setlength{\parindent}{0pt}
Humanoid controllers have acquired increasingly diverse motor skills, yet translating these capabilities into precise and stable whole-body manipulation remains challenging. Manipulation tasks are sensitive to execution deviations -- even small errors or unintended steps can cause failure. 
Introducing \modelname, a whole-body controller for precise and stable whole-body manipulation, together with a benchmark spanning diverse coordination demands. 
\modelname adopts calibrated sparse motion references to preserve task-relevant human motion structure without online joint-space retargeting, leaving kinematic freedom for precise control. 
To guide this freedom towards stable, physically plausible whole-body coordination, we shape the training distribution through offline curation of robot motions paired with these calibrated sparse references. 
We further introduce explicit error feedback decoupled from motion encoding, improving control precision while supporting flexible command interfaces. 
To achieve a systematic evaluation of these capabilities, 
a benchmark is proposed to assess precision, sustained execution stability, and reachability across diverse whole-body coordination demands. Evaluations in simulation and on hardware show stronger precision and stability across all evaluation tracks. 
Under the most strict real-world global scoring setting, \modelname achieves a 30\% to 80\% decrease in tracking error with fewer look-ahead inputs, 
and without retargeting compared to all existing controllers. 
Downstream experiments demonstrate more efficient teleoperated data collection and higher task success using the same deployment pipeline as well. 
We believe Anchor 1 supports precise and stable whole-body control as a practical basis for humanoid task execution.
\end{abstract}

\section{Introduction}
\label{sec1:intro}

Humanoid robots offer the prospect of performing everyday tasks in environments designed around the human body~\cite{gu2025humanoid}.
Realizing this prospect requires accurate coordination of the hands, trunk, and feet as the robot moves and interacts with its surroundings.
Reaching into a low shelf couples hand placement with changes in posture and balance; carrying a tray requires its pose to be maintained while the feet reposition.
In both cases, small deviations at the end effectors can disrupt the task even when the robot remains upright, and an unintended step can shift the body and displace the hands from their targets.
A controller for whole-body manipulation must therefore be precise, bringing the hands and feet to their intended targets, and stable, sustaining this accuracy throughout execution without falls, slips, or unintended steps.
These capabilities underpin predictable teleoperation, successful demonstration collection~\cite{He2024omniho,Fu2024humanplus,Ze2025TWIST2}, and reliable execution by autonomous policies~\cite{jiang2025wholebodyvla}.

Learning from human motion have enabled substantial progress in whole-body control~\cite{Cheng2024exbody,Chen2025gmt,Qi2026HumanoidGPT}.
Recent work demonstrates agile, natural movement and transfer across motion sources and embodiments~\cite{liao2026beyondmimic,sleiman2026zest,Liao2026SuperMimic}, alongside general-purpose interfaces for interactive and autonomous control~\cite{luo2026sonic,Zeng2026ScaleBFM,Li2026BFMZero}.
Task-oriented systems further demonstrate coordinated whole-body control for dynamic interactions~\cite{Jiang2026PingpongDuo,Ren2026SMASH,Zhong2026RoboNaldo}.
In parallel, other approaches transfer egocentric human demonstrations to humanoid loco-manipulation through visual and action alignment~\cite{shi2026egohumanoid,Chen2026EgoHumanoidV2}.
Bringing these advances to whole-body manipulation raises two fundamental requirements: task intent must be faithfully expressed in a control command, and this command must be realized through precise, stable execution.
In teleoperation, the command must adapt the operator's motion to the robot's morphology while preserving task-relevant cues and a predictable correspondence.
A common approach constructs the command as a robot-space reference through online retargeting~\cite{Araujo2026GMR,Cao2026UMR}; errors in this conversion can disrupt the intended correspondence or introduce discontinuities.
Encoding human motion directly avoids this conversion~\cite{luo2026sonic}, but a parametric body representation~\cite{SMPL:2015} that disregards the operator's body proportions can distort task-relevant trajectory structure.

Even with a faithful command, close reproduction of the overall motion does not ensure sustained accuracy at task-relevant targets.
The controller needs to correct deviations as posture and support change, without introducing unintended movements that disrupt the task.
Existing controllers obtain the signal for this correction in different ways.
Token-conditioned controllers~\cite{luo2026sonic} receive the intended motion as a learned token together with proprioception. This compact, structured representation can facilitate learning by higher-level policies. However, tracking-error feedback is implicit in this formulation: the policy must infer deviations from the token and proprioception, while compression obscures fine-grained target information and makes small residual errors harder to correct. 
By contrast, explicit tracking-error feedback provides a direct correction signal~\cite{He2024omniho,Pan2025AMS,Zeng2026ScaleBFM}, but encoding it in the motion token would entangle intended behavior with transient execution errors, complicating token prediction and reuse by higher-level policies.

Progress toward such controllers also depends on evaluations that can verify these capabilities.
Such evaluations must distinguish overall motion reproduction from sustained accuracy under manipulation-relevant conditions.
Humanoid benchmarks now span whole-body task learning and the interplay between high-level policies and low-level trackers~\cite{Sferrazza2024HumanoidBench,Wang2026HumanoidArena}.
Motion-tracking benchmarks have also broadened motion coverage and begun to assess contact quality and alignment with human preferences~\cite{motionchallenge2026,liu2026humantrackercomprehensivehumanalignedmotion}.
Yet task success in task-level benchmarks also reflects perception and planning, whereas sequence-averaged tracking errors can obscure brief deviations.
Complementary, controller-level evaluations are therefore needed to characterize how precisely targets can be reached, how consistently they can be maintained, and under which whole-body conditions execution becomes unreliable.

Here, we present \modelname, a humanoid controller for precise and stable whole-body manipulation.
\modelname constructs its commands through a structure-preserving calibration of human motion to robot scale, which retains task-relevant motion cues and a predictable correspondence without online joint-space retargeting.
To focus control on the body parts that most directly shape manipulation and support, we restrict these commands to sparse motion references~\cite{tessler2024maskedmimic,he2025hover,Zeng2026ScaleBFM} that specify target poses only for the pelvis, both wrists, and both ankles.
With only these poses specified, the policy can use the remaining kinematic redundancy to track them precisely.
This freedom, however, also admits multiple whole-body realizations, leaving the training distribution to determine how it is used.
We shape this distribution by pairing each human motion with an offline-retargeted robot reference and curating these pairs through quality screening.
The resulting data bias the learned coordination toward stable, physically plausible behavior.

To retain explicit error feedback for precise tracking while supporting flexible commands from higher-level policies, \modelname accepts either sparse motion references or motion tokens~\cite{luo2026sonic} and decouples this feedback from motion encoding.
Execution errors thus guide control without entering the motion tokens.
When commands arrive as tokens, targets reconstructed from them provide the reference for this feedback.
The reconstruction objective also ties each token to the intended motion, keeping the token space consistent for higher-level policies to learn.

To fairly assess these control capabilities of current whole-body controllers, we also introduce a benchmark across three complementary tracks, especially highlighting out-of-distribution whole-body manipulation-related motions.
The In-house track uses task-structured motion-capture sequences that combine upper-body actions with varied lower-body support and movement, probing coordination during reaching and hand stabilization.
The Public track broadens the evaluation to diverse motion sources and behaviors, including dynamic whole-body movements.
The Reachability track tests whether prescribed wrist targets can be reached and held across body configurations without foot slips or extra steps.
Together, these tracks characterize sustained tracking accuracy, reachability, and stability under varied whole-body coordination demands.

Our contributions are threefold.
First, we present \modelname, a whole-body controller that integrates structure-preserving sparse references, stability-oriented distribution curation, and decoupled error feedback for precise and stable whole-body manipulation.
Second, we introduce a controller-level benchmark for diagnosing precision and stability under diverse conditions relevant to whole-body manipulation.
Third, through simulation and hardware evaluations, we show that \modelname improves tracking precision and execution stability over the evaluated baselines and yields more efficient teleoperated data collection and higher autonomous task success.
Hardware demonstrations further cover fine end-effector contact and fast interception, as well as whole-body manipulation while kneeling, bending, or walking (\cref{fig:teaser}).

\section{Results}
\label{sec2:results}

\begin{figure}[t]
    \centering
    \includegraphics[width=\linewidth,height=0.44\textheight,keepaspectratio]{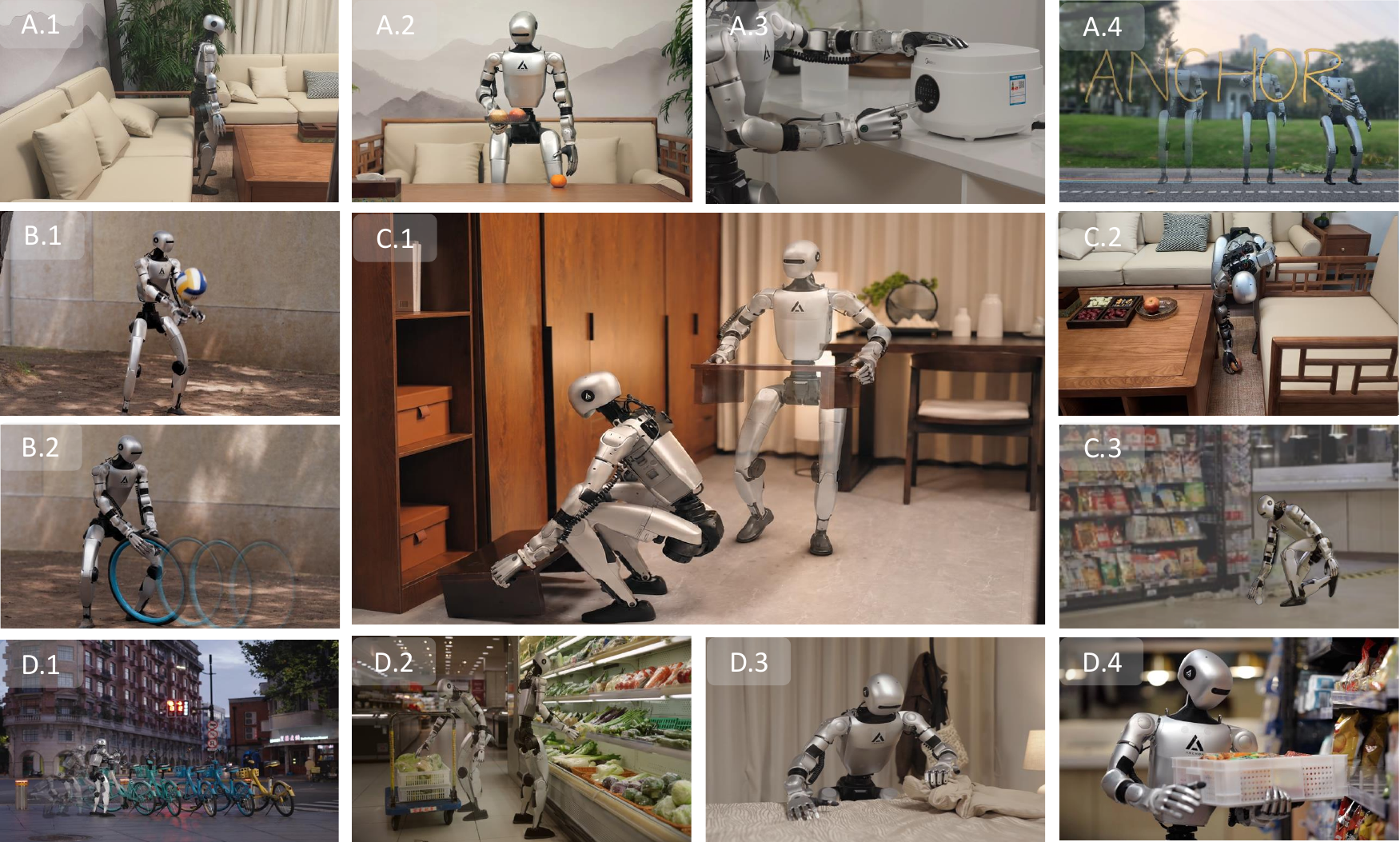}
    \captionsetup{justification=raggedright,singlelinecheck=false}
    \caption{\textbf{\modelname -- Precise, responsive, and stable whole-body manipulation.}
    \textbf{(A) Precision.} A.1 navigating a narrow passage; A.2 carrying a plate through a narrow passage; A.3 pressing a rice cooker button; and A.4 tracing characters with the end effector.
    \textbf{(B) Low-latency response.} B.1 accurately striking a fast-moving volleyball; and B.2 rapidly and precisely intercepting a narrow-profile tire.
    \textbf{(C) Postural stability.} C.1 holding a wooden stool and place it on the floor. C.2 bending to retrieve an orange in a narrow space; and C.3 picking up a water bottle while stably kneeling on one knee.
    \textbf{(D) Whole-body coordination.} D.1 lifting a bicycle upright; D.2 pulling a grocery cart while placing groceries at designated shelf locations; D.3 bending to arrange bedding; and D.4 steadying a basket with one hand while stacking goods with the other during locomotion.}
    \label{fig:teaser}
\end{figure}

\modelname supports accurate reaching, object-pose maintenance during body motion, and coordinated whole-body manipulation, as illustrated by the real-world demonstrations in \cref{fig:teaser}.

A structured benchmark measures tracking precision, stability, and reachability through comparisons in simulation and on hardware (\cref{sec2:subsec:benchmark,sec2:subsec:sim,sec2:subsec:real}).
Ablation studies and supporting experiments examine the reference representation and controller architecture (\cref{sec2:subsec:validation}), followed by evaluations of teleoperated data collection and autonomous task execution (\cref{sec2:subsec:teleop-vla}).

\FloatBarrier
\begingroup
\renewcommand{\topfraction}{0.9}
\renewcommand{\bottomfraction}{0.8}
\renewcommand{\textfraction}{0.1}
\subsection{Benchmarking manipulation-centric whole-body control}
\label{sec2:subsec:benchmark}

Benchmarking whole-body manipulation requires evaluating positioning accuracy and support stability alongside general motion-tracking performance.
The benchmark therefore comprises three tracks: In-house, Public, and Reachability.
The In-house track systematically varies the coordination demands placed on the controller through structured motion-capture tasks.
The Public track extends evaluation to external motion sources and behaviors beyond those emphasized by the in-house tasks.
The Reachability track adds prescribed spatial targets and holding phases, allowing positioning performance to be examined alongside support stability.
Together, these tracks assess controller performance across both diverse motion sequences and explicitly specified reaching conditions.

\subsubsection{Benchmark composition and motion diversity}
\label{sec2:subsec:benchmark:subsubsec:data}

The \textbf{In-house track} contains 15,342 motion-capture sequences across 3,247 tasks designed to probe manipulation-related whole-body coordination.
We specify torso configuration and arm-and-hand behavior separately from lower-body support and movement, then combine compatible action descriptions into motion-capture instructions.
These tasks include unilateral and bimanual reaching from different postures, coordinated hand--foot motion, and hand stabilization during body movement.
The task structure makes the intended body behavior and its lower-body context explicit, allowing related manipulation demands to be examined under different postural and locomotor conditions.
The associated tags also allow results to be grouped by coordination demand.
Tag coverage, action-tag distributions, and details of the spatial analysis are provided in \cref{supp:sec2:coverage}.

The \textbf{Public track} contains 2,230 motion sequences selected from BONES-SEED~\cite{bones_seed_dataset}, CIRCLE~\cite{araujo2023circle}, OMOMO~\cite{li2023object}, and AMASS~\cite{mahmood2019amass}.
We select BONES-SEED and CIRCLE sequences by actor or subject identity, use the existing OMOMO test partition, and include a predefined AMASS test subset.
These sequences introduce motion patterns from external datasets, including running and jumping, which are absent from the In-house track.
The Public track therefore complements the manipulation-focused task composition of the In-house track by testing whether tracking capability extends to a wider range of motion sources and dynamic behaviors.

The \textbf{Reachability track} complements these motion-sequence evaluations with explicitly specified Cartesian wrist targets.
We define fixed right-wrist targets in a world coordinate frame across five postures: standing, shallow squat, deep squat, staggered stance, and half kneeling.
Each case pairs a target location with a posture.
We generate base motions for each posture using Kimodo~\cite{rempe2026kimodo}, then adapt them to the targets through constrained inverse kinematics subject to joint limits and prescribed support contacts.
The resulting references include preparation, reaching, target holding, and return phases.
Accepted references maintain a maximum wrist-position error below 1~cm relative to the prescribed target during the holding interval.
After quality control and left--right mirroring, the track contains 3,210 reference motions covering reaching with either wrist.
This construction provides target-conditioned references for evaluating sustained tracking across body configurations.
Tracking success, foot-slip and extra-step events are recorded over the reach--hold--return task window, testing whether the controller can execute the motion while maintaining the intended support.

The coverage analyses further illustrate the complementary composition of the tracks.
Joint projections of semantic and pose features provide a qualitative view of overlapping and distinct motion regions across all three tracks (\cref{fig:bench-overview}B).
Under the coordinate transformations used for the spatial analysis, the In-house references exhibit broader wrist-position coverage, particularly in the stationary subset, whereas the Public references span a larger region of pelvis displacement after initial alignment (\cref{fig:bench-overview}C).
These distributions characterize the spatial occupancy of the reference motions; the Reachability track separately specifies target locations across postures for controller assessment.

\begin{figure}[t]
    \centering
    \includegraphics[width=\linewidth,height=0.80\textheight,keepaspectratio]{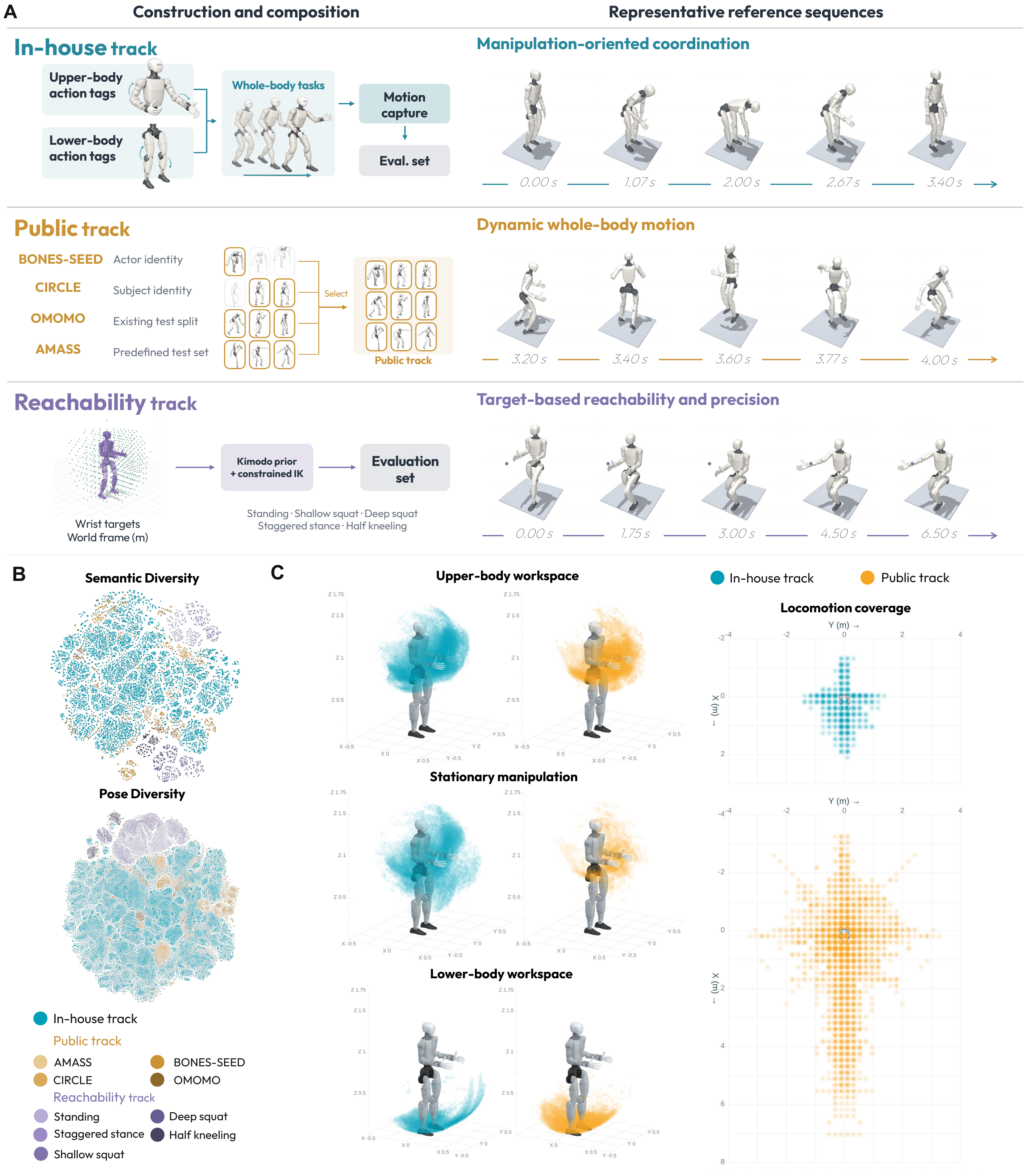}
    \caption{\textbf{Benchmark construction and motion coverage.}
    \textbf{(A)} Construction of the In-house, Public, and Reachability tracks, with representative reference sequences and timestamps in seconds; purple markers indicate wrist targets.
    \textbf{(B)} Joint t-SNE projections of all three tracks, providing a qualitative view of their extensive semantic and pose diversity.
    \textbf{(C)} Spatial distributions of wrist and ankle positions and pelvis trajectories for the In-house and Public tracks, showing upper and lower-body workspaces, stationary manipulation, and locomotion coverage.
    The two tracks provide complementary coverage of manipulation-related body configurations and global mobility.}
    \label{fig:bench-overview}
\end{figure}

Feature-space distributions and spatial coverage of the three tracks are visualized in \cref{fig:bench-overview}B and C. All benchmark sequences are held out from training.

\FloatBarrier
\subsubsection{Evaluation criteria}
\label{sec2:subsec:benchmark:subsubsec:metrics}

Benchmarking whole-body manipulation requires evaluation criteria that capture both tracking precision and the ability to sustain it throughout execution.
Tracking precision is quantified by mean position errors, and sustained tracking by motion-level success rates (SR).
Global scoring compares executed and reference trajectories after a fixed initial alignment of horizontal pelvis position and heading.
Local scoring additionally removes each trajectory's instantaneous horizontal pelvis translation while preserving height and heading differences.
Both scoring frames are independent of the controller's input mode.

For the In-house and Public tracks, a motion is successful only if wrist and ankle position errors remain within 10~cm, their orientation errors within 1~rad, and all other tracked body-point position errors within 25~cm throughout the sequence.
All motions, including failures, contribute equally to SR.
Mean position errors are averaged within each motion and then across motions, using frames preceding the first position error exceeding 25~cm at any tracked body point.
The resulting observation windows may differ across controllers.

For the Reachability track, task success requires global tracking without foot slips or extra steps throughout the reach--hold--return task window.
Mean slip and extra-step counts per motion further characterize support stability.
Full definitions, failure handling, and aggregation rules are provided in \cref{sec4:subsec:evaluation:subsubsec:metrics}.

\FloatBarrier
\endgroup
\subsection{Tracking precision, stability, and reachability in simulation}
\label{sec2:subsec:sim}

\begin{figure}[t]
    \centering
    \includegraphics[width=\linewidth,height=0.70\textheight,keepaspectratio]{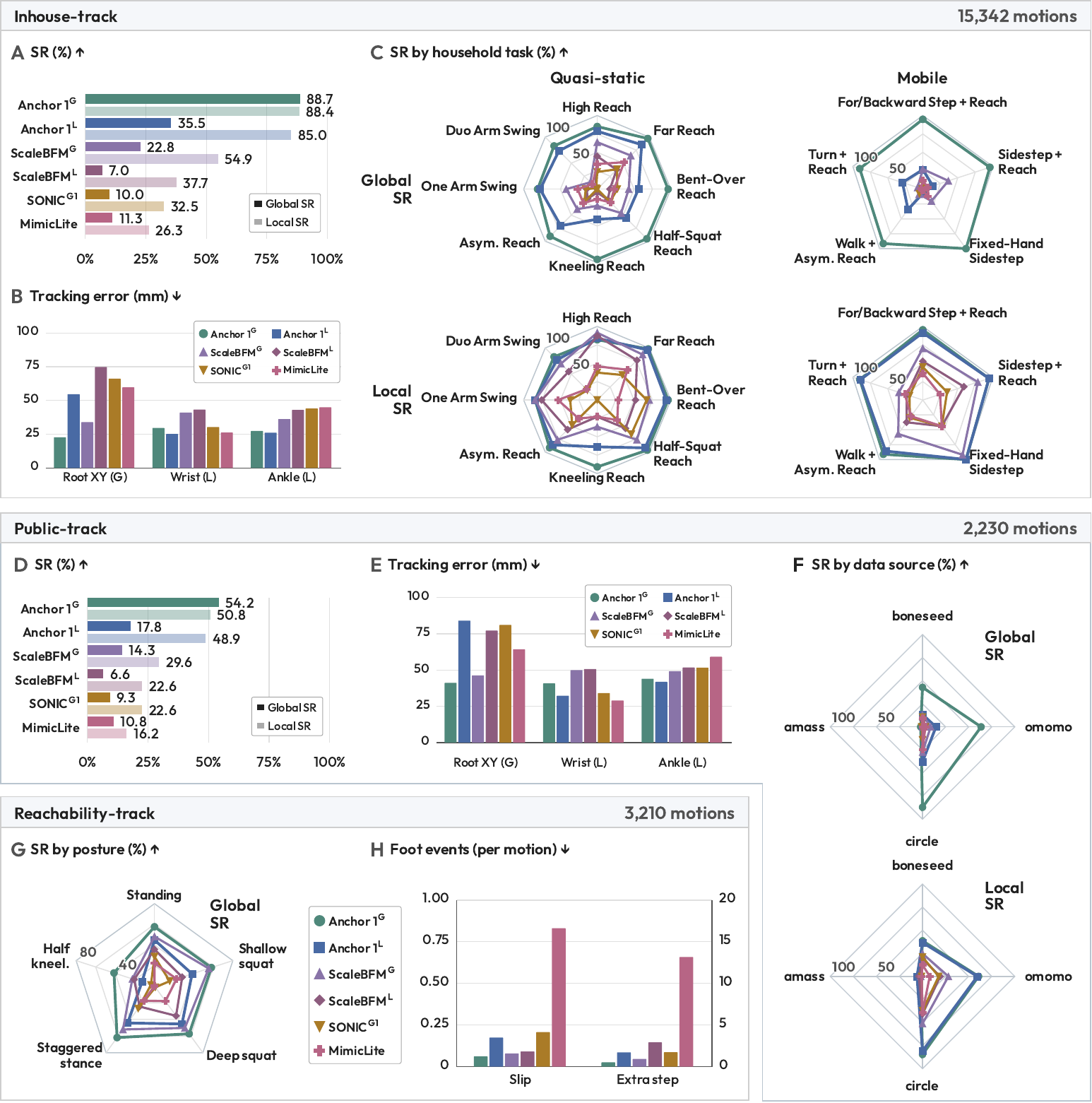}
    \caption{\textbf{Tracking accuracy and reachability across three tracks.}
    All results are evaluated in MuJoCo.
    \textbf{(A, D)} Local and global success rates (SR) on Inhouse-track (15,342 motions) and Public-track (2,230 motions).
    \textbf{(B, E)} Mean position errors in millimeters for horizontal root position (global), wrists, and ankles (local).
    \textbf{(C)} Global (top) and local (bottom) SR across quasi-static and mobile household task categories in Inhouse-track.
    Categories include complete motions containing the indicated phases and may overlap.
    \textbf{(F)} Global (top) and local (bottom) SR by data source in Public-track.
    \textbf{(G)} Global task SR on 3,210 Reachability-track references, grouped by posture.
    Left and right variants are pooled for posture-wise SR.
    \textbf{(H)} Mean slips (left axis) and extra steps (right axis) per Reachability-track motion.
    Reachability metrics use the evaluation window defined in \cref{sec4:subsec:benchmark:subsubsec:reachability}.
    Superscript $G$ denotes global mode for both \modelname and ScaleBFM.
    Superscript $L$ denotes re-locate mode for \modelname and reference-forcing mode for ScaleBFM.
    These mode labels are independent of the local and global scoring frames.
    SONIC's G1 superscript identifies its robot encoder; Asym. denotes asymmetric.
    Metric definitions and aggregation are given in \cref{sec4:subsec:evaluation:subsubsec:metrics}.}
    \label{fig:sim-results}
\end{figure}

Simulation experiments compare \modelname with ScaleBFM XL~\cite{Zeng2026ScaleBFM}, SONIC v1.1~\cite{luo2026sonic}, and MimicLite v1.1~\cite{RoboParty2026MimicLite} to assess tracking precision, sustained tracking success, and reachability across the three benchmark tracks (\cref{fig:sim-results}).
Within each track, all configurations are tested on the same reference motions using the criteria in \cref{sec2:subsec:benchmark:subsubsec:metrics} and a common simulation and domain-randomization protocol (\cref{sec4:subsec:evaluation:subsubsec:simulation}).
References are precomputed for all tracks, with sparse robot-space references supplied to \modelname on the Reachability track.
The comparisons assess complete controller configurations. 
Controller look-ahead and online retargeting requirements are summarized in \Cref{tab:real-main}.
Corresponding results in Isaac Sim are provided in \cref{supp:sec3:isaacsim}.

On the In-house track, \modelname in global mode achieves the highest local and global success rates (SR), at 88.4\% and 88.7\%, respectively, and the lowest mean horizontal root position error among the compared configurations, alongside low wrist and ankle position errors (\cref{fig:sim-results}A,B).
The category-wise comparisons also show higher tracking success on household motions requiring whole-body coordination, including low-reaching and half-kneeling reaches (\cref{fig:sim-results}C).
This advantage extends to manipulation during locomotion: during walking with asymmetric bimanual reaches, its local SR is 91.2\%, compared with 57.0\% for ScaleBFM in global mode, the strongest baseline in this setting.

The Public track extends evaluation to highly dynamic behaviors, including running and jumping, that are absent from the In-house track and less represented in training.
The three baseline methods may have been exposed to portions of the public datasets during training.
Although absolute success rates are lower on this track, \modelname in global mode retains the highest local and global SR among the compared configurations (\cref{fig:sim-results}D).
It also has the lowest horizontal root position error and lower ankle position error than all baselines (\cref{fig:sim-results}E).
Success rates are particularly low across controllers on the AMASS test subset, indicating that its motions remain challenging under the sustained-tracking criterion (\cref{fig:sim-results}F).
These results support whole-body tracking capability beyond the behaviors emphasized by the In-house track.

Across the In-house and Public tracks, \modelname in re-locate mode retains local tracking performance close to that of global mode, with similar pelvis-relative wrist and ankle position errors (\cref{fig:sim-results}A,B,D,E).
The preservation of local tracking accuracy supports its use in teleoperation, where re-locating references to the robot's current root accommodates differences between the operator's and robot's global trajectories.

On the Reachability track, \modelname in global mode achieves the highest global task SR in most posture groups, including standing, squatting, and staggered stance (\cref{fig:sim-results}G).
Mean foot-event counts are low, at 0.06 slips and 0.53 extra steps per motion (\cref{fig:sim-results}H).

\clearpage
\subsection{Tracking accuracy and task success on the real robot}
\label{sec2:subsec:real}

Real-world evaluation covers two complementary motion subsets drawn from the benchmark tracks (\cref{sec2:subsec:benchmark:subsubsec:data}).
The Reachability subset comprises standing reaches selected to broadly cover the workspace and assess end-effector tracking precision.
The In-house subset contains motions selected using predefined action tags to examine performance across structured task categories.

\modelname is compared with MimicLite v1.1~\cite{RoboParty2026MimicLite}, the 5-point and 14-point configurations of ScaleBFM~\cite{Zeng2026ScaleBFM}, and SONIC v1.1~\cite{luo2026sonic} with G1 and human-body inputs.
Both subsets report mean wrist and ankle position and orientation errors in global and local scoring frames.
Body-point poses are reconstructed through forward kinematics from recorded joint configurations and motion-capture measurements of the pelvis pose.
The error statistics include all valid recorded samples, without the 25~cm position-error truncation applied in simulation.

Global tracking error measures world-space deviations from the reference trajectories after a fixed initial alignment of horizontal pelvis position and heading.
Local position error reduces the contribution of pelvis translation, providing a complementary assessment of limb tracking.
Low local error alone does not establish accurate global execution, because some body-displacement errors are removed by the local scoring frame.
Reporting both therefore distinguishes local tracking precision from accuracy along the world-space trajectory.

The In-house subset additionally reports motion-level success rate (SR), following the simulation success criteria.
A motion succeeds only if the prescribed body-point position and end-effector orientation thresholds are satisfied throughout its valid recorded segment.
SR is calculated within each task category as the fraction of successful motions among those with valid recordings.

\begin{table}[t]
\centering
\caption{\textbf{Real-world tracking errors and reference requirements across subsets of two tracks.}
Reachability comprises 51 standing reaches selected to broadly cover the workspace; In-house comprises 72 tag-selected whole-body manipulation motions.
Both subsets report global and local wrist and ankle position and orientation errors.
Reachability uses the right wrist and bilateral ankle means; in-house uses bilateral means for both wrists and ankles.
$p$ and $r$ denote mean position error (mm) and mean orientation error (degrees), respectively.
Subscripts $w$ and $a$ identify wrists and ankles; superscripts $G$ and $L$ denote global and local scoring frames, respectively.
$N$ denotes the target motion count for each subset.
Look-ahead is the maximum future-reference horizon in seconds, excluding historical inputs.
Input identifies the reference format ($R$, $H$, or $H^{\prime}_{5}$).
The control requirements apply to both subsets.
\cmark\ indicates that online retargeting is not required; \xmark\ indicates that it is required.
Lower errors are better; bold values indicate the lowest error in each metric within each subset.}
\label{tab:real-main}
\fontsize{9.5}{11.5}\selectfont
\setlength{\tabcolsep}{2pt}
\renewcommand{\arraystretch}{1.15}
\begin{tabular*}{\linewidth}{@{\extracolsep{\fill}}lrcc*{8}{r}@{}}
\toprule
& \multicolumn{3}{c}{\textbf{Control requirements}}
& \multicolumn{4}{c}{\textbf{Wrist}}
& \multicolumn{4}{c}{\textbf{Ankle}} \\
\cmidrule(lr){2-4}\cmidrule(lr){5-8}\cmidrule(lr){9-12}
\textbf{Method}
& \shortstack{\textbf{Look-ahead}\\(s)}
& \shortstack{\textbf{No}\\\textbf{retargeting}}
& \textbf{Input}
& \shortstack{$p_w^G$\\(mm)}
& \shortstack{$r_w^G$\\($^\circ$)}
& \shortstack{$p_w^L$\\(mm)}
& \shortstack{$r_w^L$\\($^\circ$)}
& \shortstack{$p_a^G$\\(mm)}
& \shortstack{$r_a^G$\\($^\circ$)}
& \shortstack{$p_a^L$\\(mm)}
& \shortstack{$r_a^L$\\($^\circ$)} \\
\midrule
\multicolumn{12}{l}{\textbf{(a) Reachability} ($N=51$)} \\
\addlinespace[2pt]
MimicLite~\cite{RoboParty2026MimicLite}
& 0.08 & \xmark & $R$
& 53.59 & 5.56 & 27.87 & 8.22
& 50.40 & 6.58 & 46.15 & 6.58 \\
ScaleBFM-14~\cite{Zeng2026ScaleBFM}
& 0.10 & \xmark & $R$
& 64.18 & 6.40 & 46.15 & 8.26
& 66.05 & 8.34 & 65.75 & 8.34 \\
ScaleBFM-5~\cite{Zeng2026ScaleBFM}
& 0.10 & \xmark & $R$
& 72.69 & 8.34 & 51.22 & 8.25
& 77.13 & 9.68 & 74.64 & 9.68 \\
SONIC-G1~\cite{luo2026sonic}
& 0.90 & \xmark & $R$
& 98.15 & 10.19 & 42.38 & 9.42
& 92.46 & 9.45 & 69.49 & 9.45 \\
SONIC-SMPL~\cite{luo2026sonic}
& 0.18 & \cmark & $H$
& 166.70 & 19.20 & 85.87 & 18.75
& 134.30 & 12.48 & 71.93 & 12.48 \\
\addlinespace[3pt]
\textbf{\modelname}
& 0.06 & \cmark & $H^{\prime}_{5}$
& \textbf{44.58} & \textbf{4.67}
& \textbf{24.72} & \textbf{5.12}
& \textbf{35.70} & \textbf{5.76}
& \textbf{35.13} & \textbf{5.76} \\
\midrule
\multicolumn{12}{l}{\textbf{(b) In-house} ($N=72$)} \\
\addlinespace[2pt]
MimicLite~\cite{RoboParty2026MimicLite}
& 0.08 & \xmark & $R$
& 105.74 & 5.16 & \textbf{25.63} & 5.16
& 115.03 & 5.96 & 52.06 & 5.96 \\
ScaleBFM-14~\cite{Zeng2026ScaleBFM}
& 0.10 & \xmark & $R$
& 218.62 & 7.82 & 67.71 & 7.82
& 213.34 & 6.43 & 68.98 & 6.43 \\
ScaleBFM-5~\cite{Zeng2026ScaleBFM}
& 0.10 & \xmark & $R$
& 209.15 & 6.97 & 63.85 & 6.97
& 213.21 & 6.43 & 76.23 & 6.43 \\
SONIC-G1~\cite{luo2026sonic}
& 0.90 & \xmark & $R$
& 234.37 & 9.34 & 59.36 & 9.34
& 245.01 & 8.53 & 83.16 & 8.53 \\
SONIC-SMPL~\cite{luo2026sonic}
& 0.18 & \cmark & $H$
& 314.00 & 24.89 & 114.04 & 24.89
& 297.91 & 11.01 & 97.70 & 11.01 \\
\addlinespace[3pt]
\textbf{\modelname}
& 0.06 & \cmark & $H^{\prime}_{5}$
& \textbf{75.69} & \textbf{4.47}
& 29.43 & \textbf{4.47}
& \textbf{70.52} & \textbf{4.91}
& \textbf{27.96} & \textbf{4.91} \\
\bottomrule
\end{tabular*}
\end{table}

\subsubsection{Tracking precision on the Reachability subset}
\label{sec2:subsec:real:subsubsec:accuracy}
The Reachability subset comprises 51 standing-reach motions selected to broadly cover the wrist-target workspace and this evaluation focuses on reaching precision under stationary support.

\modelname achieves the lowest global and local position and orientation errors for the right wrist and both ankles.(\cref{tab:real-main})
Its global wrist and ankle position errors are 44.58~mm and 35.70~mm, respectively, compared with 53.59~mm and 50.40~mm for the strongest baseline, MimicLite v1.1.
This improvement comes with a shorter reference look-ahead (0.06~s versus 0.08~s) and a different input requirement: \modelname accepts calibrated sparse human references $H^{\prime}_{5}$ directly, whereas MimicLite consumes robot motion references $R$, requiring online retargeting for live human input.

The comparison with SONIC-SMPL further highlights the tracking precision achievable without online retargeting.
Although both controllers accept human-motion inputs without this conversion, \modelname reduces global wrist and ankle position errors by 73.3\% and 73.4\%, respectively, while using one-third of SONIC-SMPL's reference look-ahead.
The low global errors at both the wrist and ankles are consistent with accurate reaching while maintaining stable foot placement, supporting the stable and precise whole-body coordination required for whole-body manipulation.

\begin{figure}[t]
    \centering
    \includegraphics[width=\linewidth]{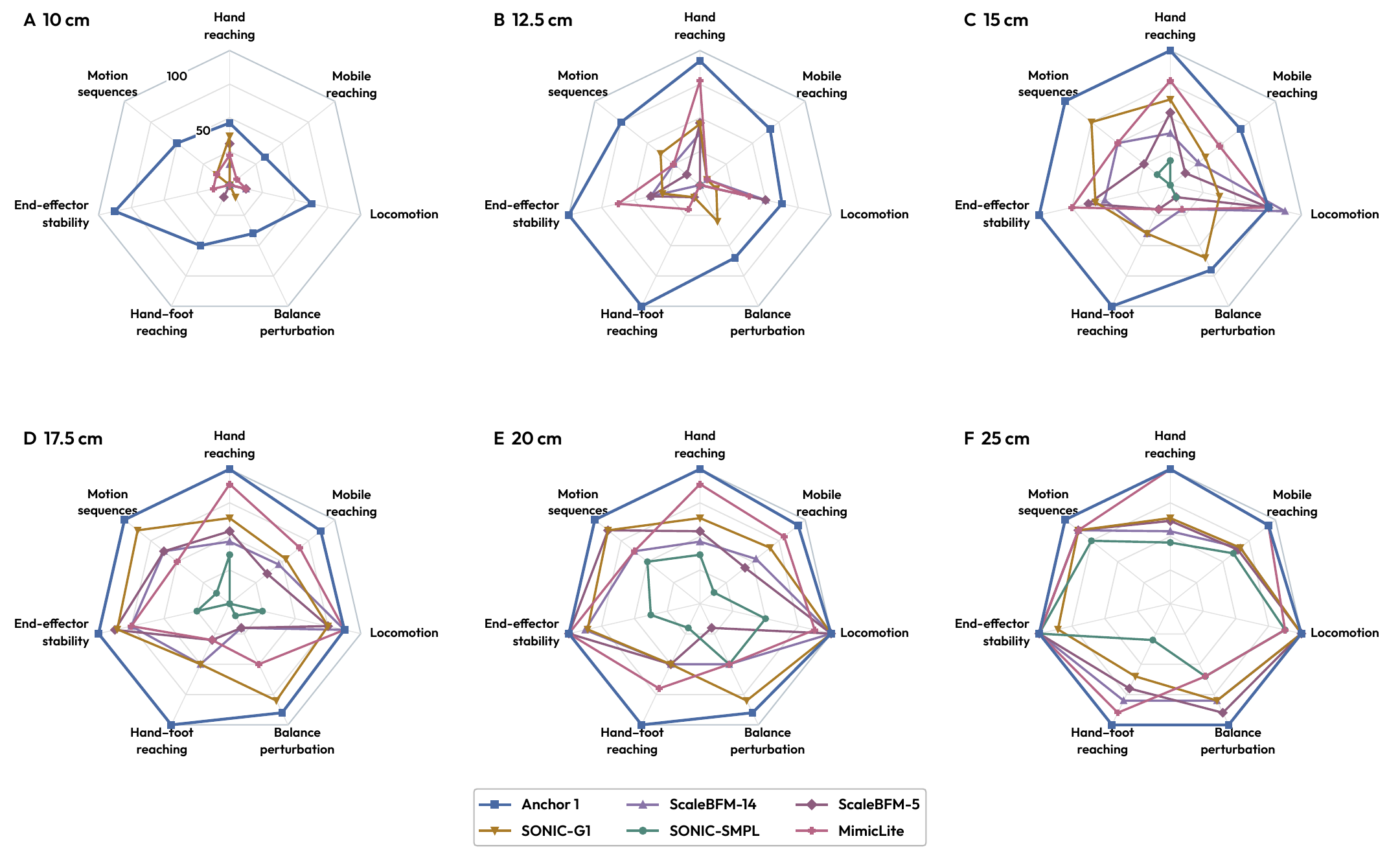}
    \caption{\textbf{Real-world task success across position thresholds on a curated subset of the In-house track.}
    Radar plots show local success rates across seven task categories for 72 motions selected by predefined action tags for real-robot testing.
    Panels \textbf{A--F} vary the wrist and ankle position tolerance from 10 to 25~cm, with all other success thresholds matching simulation (\cref{eq:tracking-sr}).
    Relaxing the tolerance narrows several category-wise differences, while \modelname maintains the highest overall success rate.}

    \label{fig:real-results}
    \label{fig:real-task-profile}
\end{figure}

\subsubsection{Task-structured evaluation on the In-house subset}
\label{sec2:subsec:real:subsubsec:success}

The In-house subset comprises 72 motions selected using predefined action tags to represent seven task categories.
These motions cover reaching, locomotion, balance, and coordinated whole-body behaviors.
Starting from the simulation-matched success criterion, the evaluation varies wrist and ankle position tolerances from 10 to 25~cm while keeping all other thresholds fixed.
\modelname achieves the highest overall success rate across all six position thresholds (\cref{fig:real-results}), leading the strongest baseline by 41.2 percentage points at the strictest tolerance.
Its success profile extends across reaching, locomotion, balance perturbation, and coordinated hand--foot motions, and the overall lead persists as the tolerance increases.
These all demonstrate broad task capability across a range of precision requirements.

The tracking errors further show that this task-level advantage is accompanied by accurate coordination of multiple end effectors (\cref{tab:real-main}).
\modelname achieves the lowest errors in seven of the eight In-house metrics, including a 46.3\% reduction in local ankle position error relative to the strongest baseline.
MimicLite v1.1 attains a lower local wrist position error, but \modelname provides more accurate ankle tracking in both scoring frames.
This broader tracking accuracy complements its higher motion-level success, which requires all four end effectors to remain within the prescribed tolerances throughout each motion.

\FloatBarrier
\Needspace{7\baselineskip}
\subsection{Validation of sparse references and architecture design}
\label{sec2:subsec:validation}

\begin{figure}[!t]
    \centering
    \includegraphics[width=\linewidth]{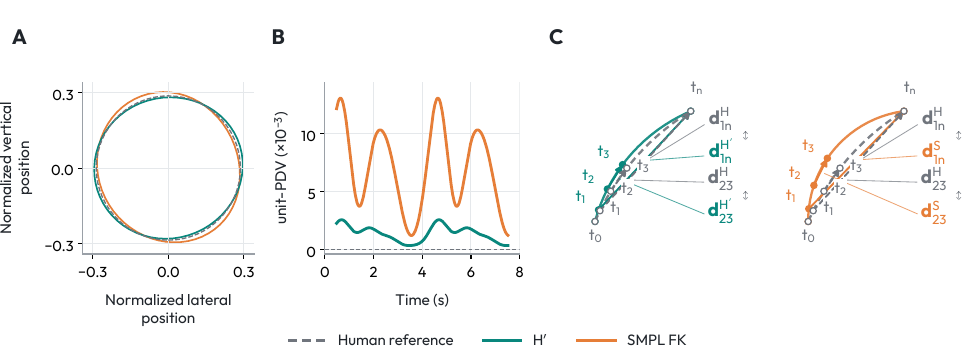}
    \caption{\textbf{Wrist reference fidelity before motion encoding.}
    \textbf{(A)} Right-wrist trajectories for a circular motion with a 16-cm human radius. Gray dashed, green, and orange curves denote the human reference, $H_5'$ targets, and SMPL forward-kinematics (FK) targets, respectively. Shoulder-relative positions are expressed in common human chest axes and normalized by each representation's fixed arm length. The $H_5'$ trajectory more closely follows the circular shape of the human reference than the SMPL FK trajectory.
    \textbf{(B)} Unit-PDV scores over time for the circular motion in (A); $H_5'$ yields an 80.2\% lower mean score than the SMPL encoder input, indicating better preservation of human wrist motion directions.
    \textbf{(C)} Schematic illustration of the unit-PDV calculation using time-matched displacement vectors to compare the human reference with $H_5'$ (left) and the SMPL encoder input (right). Calculation details are provided in \cref{supp:sec5:reference-fidelity}.}
    \label{fig:hprime-input-comparison}
\end{figure}

\subsubsection{Reference fidelity of sparse inputs}
\label{sec2:subsec:validation:subsubsec:hprime}

To preserve human motion cues before control, we evaluate the fidelity of the wrist trajectories supplied to the motion encoder.
In \modelname, these trajectories are taken from $H_5'$, a structure-preserving representation that maps human motion to robot scale through a fixed geometric calibration rather than online joint-space retargeting (\cref{sec4:subsec:formulation:subsubsec:target}).
Using the same circular human motion, we compare $H_5'$ with the input representation used by SONIC's SMPL encoder~\cite{luo2026sonic} (\cref{fig:hprime-input-comparison}).
Differences in displacement directions over matched time intervals are quantified using unit-normalized pairwise displacement-vector discrepancy (unit-PDV), with lower values indicating better preservation of directional structure (\cref{supp:sec5:reference-fidelity}).

For this motion, $H_5'$ achieves an $80.2\%$ lower mean unit-PDV than the SMPL encoder input ($1.40\times10^{-3}$ versus $7.10\times10^{-3}$).
This comparison indicates that $H_5'$ better preserves the directional structure of the human wrist trajectory.

\FloatBarrier
\begin{figure}[!t]
    \centering
    \includegraphics[width=\linewidth,height=0.70\textheight,keepaspectratio]{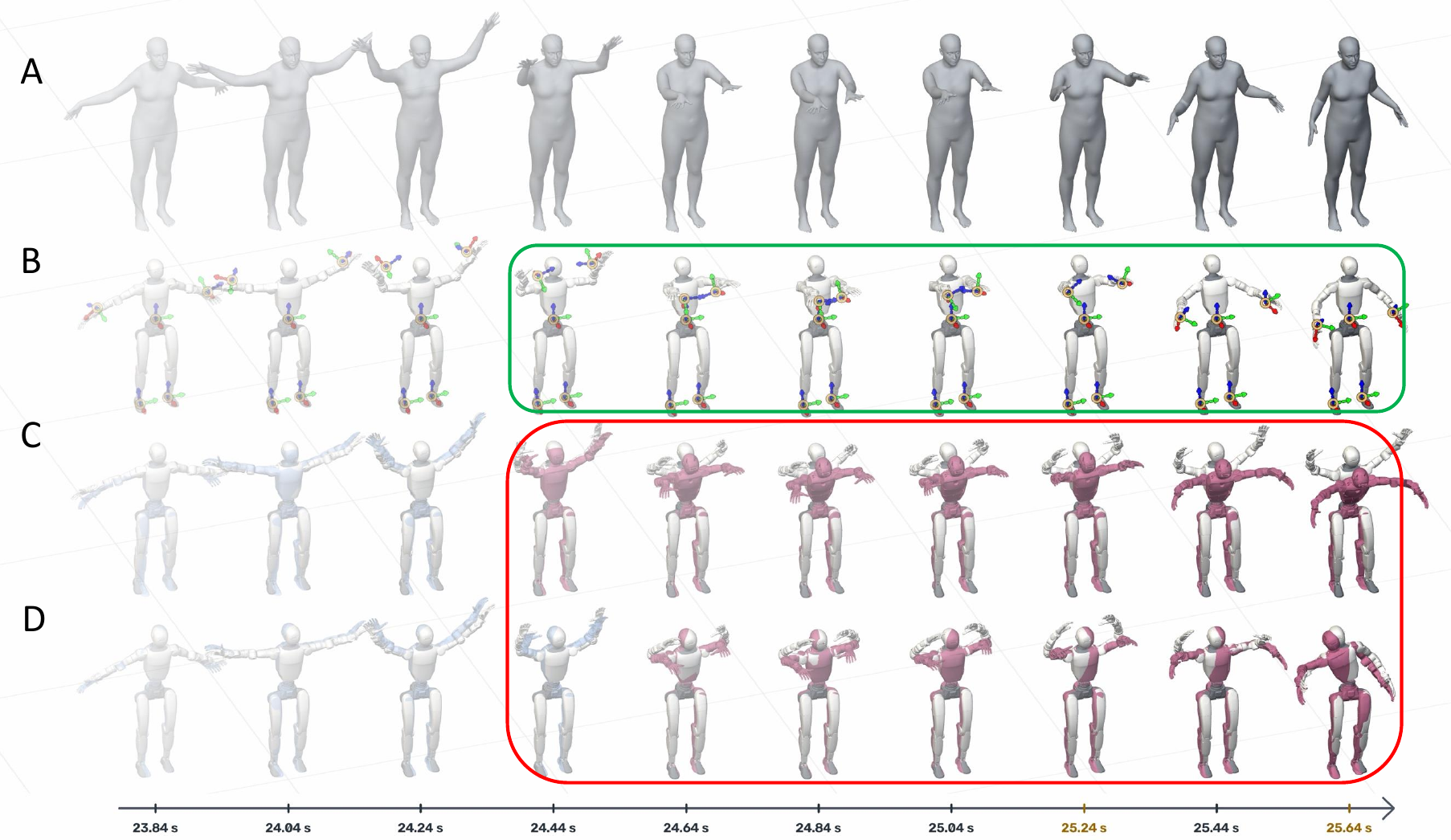}
    \caption{\textbf{Motion intent preservation and control stability under retargeting failure.}
    All configurations use the same offline human motion trajectory recorded with Pico, with time progressing from left to right.
    \textbf{(A)} Human reference motion visualized in SMPL-X format.
    \textbf{(B)} \modelname with direct sparse $H_5'$ conditioning.
    \textbf{(C)} SONIC in G1 encoder mode with robot references generated by GMR, an inverse-kinematics (IK) retargeter.
    \textbf{(D)} SONIC in G1 encoder mode with robot references generated by UMR, a learned point-cloud-matching retargeter.
    During upper-limb extension, the retargeting-dependent configurations exhibit deviations from the intended arm motion and abrupt changes, whereas \modelname retains the reference motion with stable behavior.
    Green and red outlines highlight motion preservation and retargeting-related failures, respectively.}
    \label{fig:hprime}
\end{figure}

\subsubsection{Control stability under retargeting failure}
\label{sec2:subsec:validation:subsubsec:stability}

Human-to-robot retargeting converts human motion into robot motion references before they are passed to the controller.
Errors in this conversion can produce references that deviate from the intended motion, potentially leading to unexpected behavior during execution.
This issue can arise with both inverse-kinematics (IK) solvers and learned retargeting models.
\modelname bypasses this online reference conversion by conditioning directly on $H_5'$, which is computed from the human reference to retain task-relevant spatial relationships and end-effector motion cues (\cref{sec4:subsec:formulation:subsubsec:target}).
This design aims to preserve the human motion intent without requiring a full robot joint-space reference at deployment.

In this section, we qualitatively examine this distinction by replaying the same offline human motion trajectory recorded with PICO to \modelname and SONIC~\cite{luo2026sonic} (\cref{fig:hprime}).
For \modelname, the human trajectory is converted directly into $H_5'$ and supplied to the motion encoder.
For SONIC, we use the G1 encoder to accept robot motion references produced by either General Motion Retargeting (GMR)~\cite{Araujo2026GMR}, an IK-based retargeter, or Unified Motion Retargeting (UMR)~\cite{Cao2026UMR}, a learned retargeter based on point-cloud matching.

During the illustrated upper-limb extension sequence, both retargeting-dependent SONIC configurations deviate from the intended human arm motion, whereas \modelname more faithfully preserves the reference motion cues.
The retargeted references also exhibit abrupt changes during execution, introducing discontinuous commands that can interfere with task completion and place undesirable mechanical stress on the robot joints.
These qualitative observations illustrate how retargeting errors can propagate to the controller, direct $H_5'$ conditioning avoids this intermediate source of error and maintains stable behavior while preserving the intended motion.

\FloatBarrier
\begin{table}[!t]
    \centering
    \caption{\textbf{Effects of explicit feedback, inference mode, and target reconstruction.}
    All variants use $H'_5$ as input and are evaluated in MuJoCo on the In-house track.
    \modelname\unskip$^{*}$ supplies explicit errors to the encoder; \modelname supplies them to the decoder.
    Ref. and Recon. denote supplied and reconstructed feedback targets.
    SR and Pos. err. denote success rate and mean position error over 14 body points.
    Global and local scoring frames are independent of inference mode.}
    \label{tab:ablation}
    \footnotesize
    \setlength{\tabcolsep}{3pt}
    \begin{tabular*}{\linewidth}{@{\extracolsep{\fill}}llllrrrr@{}}
        \toprule
        \raisebox{-4.5pt}[0pt][0pt]{Variant} & \multicolumn{1}{c}{Feedback} & \multicolumn{1}{c}{Inference} & \multicolumn{1}{c}{Feedback} & \multicolumn{2}{c}{Global} & \multicolumn{2}{c}{Local} \\
        & \multicolumn{1}{c}{location} & \multicolumn{1}{c}{mode} & \multicolumn{1}{c}{source} & SR (\%) $\uparrow$ & Pos. err. (mm) $\downarrow$ & SR (\%) $\uparrow$ & Pos. err. (mm) $\downarrow$ \\
        \midrule
        w/o explicit error feedback & None & Global & --- & 70.9 & 28.4 & 61.7 & 24.5 \\ %
        \modelname\unskip$^{*}$ & Encoder & Global & Ref. & 62.7 & 35.6 & 86.0 & 20.2 \\ %
        \modelname & Decoder & Global & Ref. & 88.7 & 25.5 & 88.4 & 21.9 \\ %
        \modelname & Decoder & Global & Recon. & 76.0 & 27.7 & 77.5 & 22.4 \\ %
        \modelname & Decoder & Re-locate & Ref. & 35.5 & 57.9 & 85.0 & 22.3 \\ %
        \modelname & Decoder & Re-locate & Recon. & 24.3 & 61.5 & 72.4 & 23.3 \\ %
        \bottomrule
    \end{tabular*}
\end{table}

\subsubsection{Ablation of architecture components}
\label{sec2:subsec:validation:subsubsec:ablation}
\Cref{tab:ablation} examines explicit tracking-error feedback, inference mode, and feedback target source on the In-house track in MuJoCo.
All variants retain proprioceptive observations. Explicit feedback additionally supplies wrist and ankle position and orientation errors to the encoder or decoder.
Ref. uses supplied reference poses to compute these errors, whereas Recon. uses near-term poses reconstructed from motion tokens (\cref{sec4:subsec:formulation:subsubsec:architecture}).
Global mode preserves the reference root trajectory after initial alignment, whereas Re-locate translates the reference window to align the previous reference root position with the robot's current root position (\cref{sec4:subsec:deployment}).
These inference modes are distinct from the global and local scoring frames.

With Global inference and Ref. targets, decoder feedback improves global success from 70.9\% to 88.7\% and local success from 61.7\% to 88.4\% relative to the variant without explicit feedback, while reducing position errors in both scoring frames.
Encoder feedback also increases local success from 61.7\% to 86.0\% and reduces mean local position error from 24.5 to 20.2~mm, although its global metrics are worse than those of the variant without explicit feedback.
Thus, the benefit depends on feedback placement, with decoder feedback improving both global trajectory tracking and local pose accuracy.

Re-locate retains tracking-error feedback for the translated targets but does not explicitly correct accumulated root-position drift relative to the original global trajectory.
The larger mean global position error with Ref. targets (57.9 versus 25.5~mm) is consistent with this weaker constraint on global trajectory tracking.
Local control accuracy is largely maintained, with 85.0\% success and 22.3~mm mean position error, close to the 88.4\% and 21.9~mm achieved in Global mode.

Recon. enables motion-token control without accompanying sparse references, but incurs a loss of tracking accuracy, with higher position error and lower success in both inference modes and scoring frames.
Mean local position error increases from 21.9 to 22.4~mm in Global mode and from 22.3 to 23.3~mm in Re-locate mode.
These results favor Ref. targets for precision-demanding demonstration collection when sparse references are available.
Computing feedback directly against the supplied poses allows small execution errors to be corrected without the additional approximation introduced by target reconstruction.

\FloatBarrier
\begin{figure}[!t]
    \centering
    \includegraphics[width=\linewidth,height=0.70\textheight,keepaspectratio]{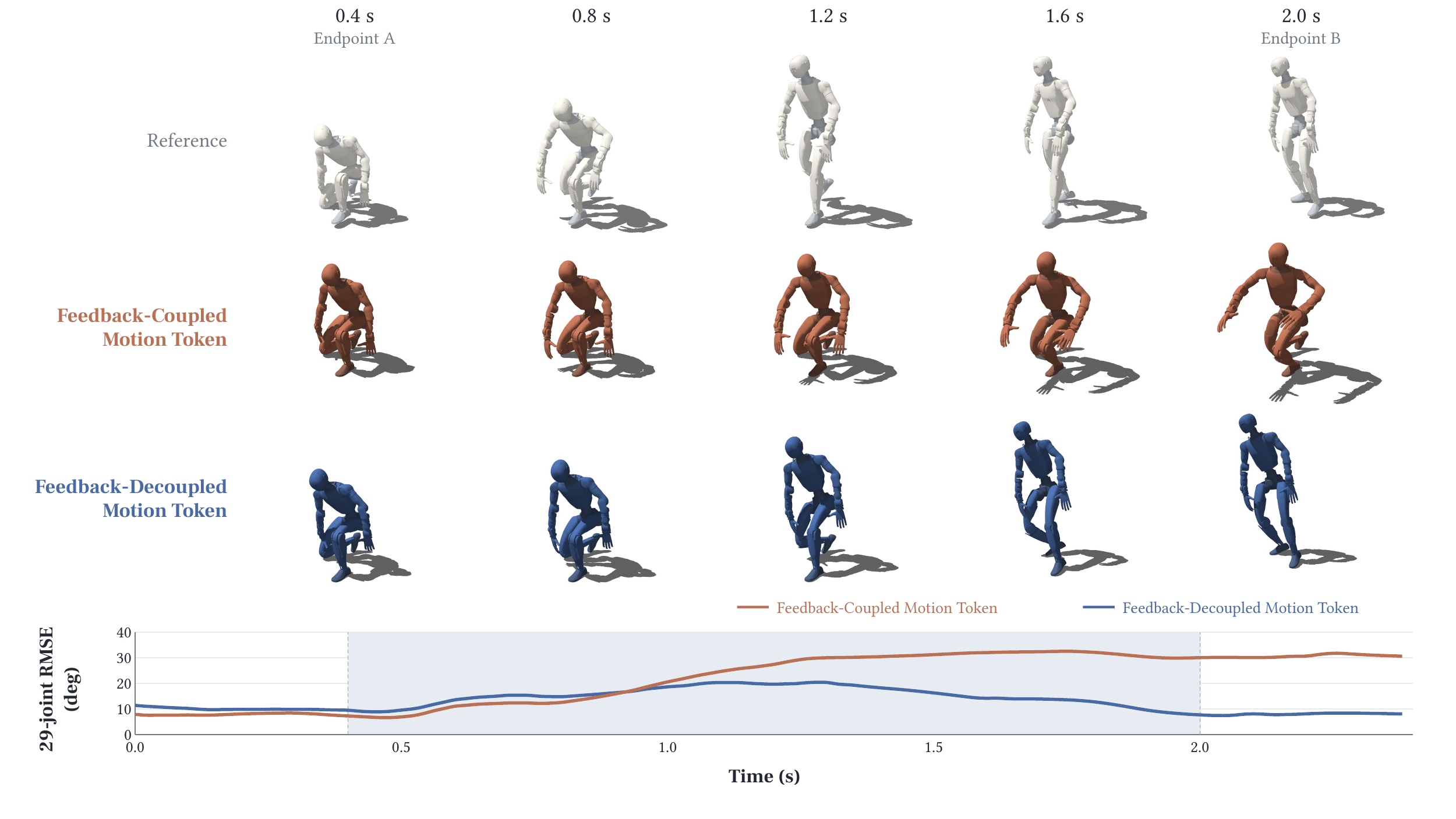}
    \caption{\textbf{Motion token interpolation with coupled and decoupled feedback.}
    Top: reference poses (gray) and robot motion under feedback-coupled (orange) and feedback-decoupled (blue) token interpolation at matching times.
    Endpoints A and B correspond to 0.4 and 2.0~s, respectively.
    The feedback-decoupled variant more closely follows the transition from a crouched posture toward the upright endpoint.
    Bottom: root mean square error (RMSE) across 29 joint angles relative to the reference, in degrees.
    The shaded region marks the interpolation interval; the feedback-decoupled variant has lower error in its later portion.}
    \label{fig:token-interp}
\end{figure}

\subsubsection{Motion token interpolation with decoupled feedback}
\label{sec2:subsec:validation:subsubsec:interpolation}

Beyond accurate tracking, a motion-token interface should support coherent robot behavior when its commands are modified or combined.
Interpolation between motion tokens provides a test of how coupling explicit tracking-error feedback with motion encoding affects this property.
We compare feedback-coupled and feedback-decoupled variants by interpolating between representations of the same two endpoint poses, spanning 0.4--2.0~s of the reference motion (\cref{fig:token-interp}).
The reference transitions from a crouched posture toward an upright posture.
The analysis includes the resulting motion snapshots and the root mean square error (RMSE) of the 29 joint angles relative to the reference.

The feedback-coupled variant remains in a more deeply flexed posture as the reference rises, whereas the feedback-decoupled variant more closely reproduces the transition toward the upright endpoint.
The error curves are similar early in the interpolation interval but diverge later, with lower reference-relative joint error for the feedback-decoupled variant toward the endpoint.
It clearly supports separating explicit execution-error correction from motion encoding to retain useful motion structure when token commands are interpolated.

\Needspace{7\baselineskip}
\subsection{Teleoperation and autonomous whole-body manipulation}
\label{sec2:subsec:teleop-vla}

\subsubsection{Teleoperation and demonstration collection}
\label{sec2:subsec:teleop-vla:subsubsec:collection}

\begin{figure}[!t]
    \centering
    {\raggedright\textbf{A}\par}
    \includegraphics[width=\linewidth]{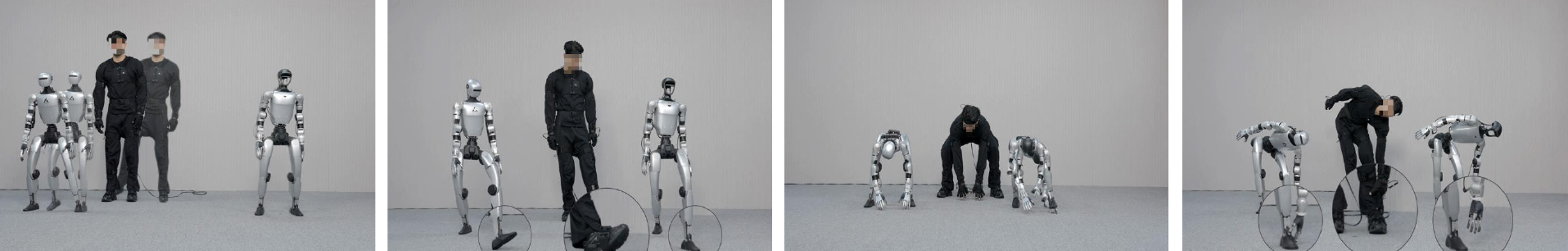}
    \par\smallskip
    {\raggedright\textbf{B}\par}
    \includegraphics[width=\linewidth]{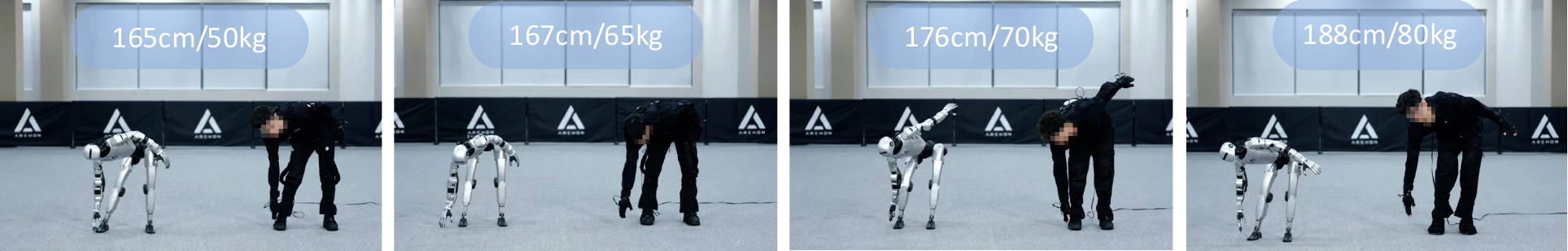}
    \par\smallskip
    {\raggedright\textbf{C}\par}
    \includegraphics[width=\linewidth]{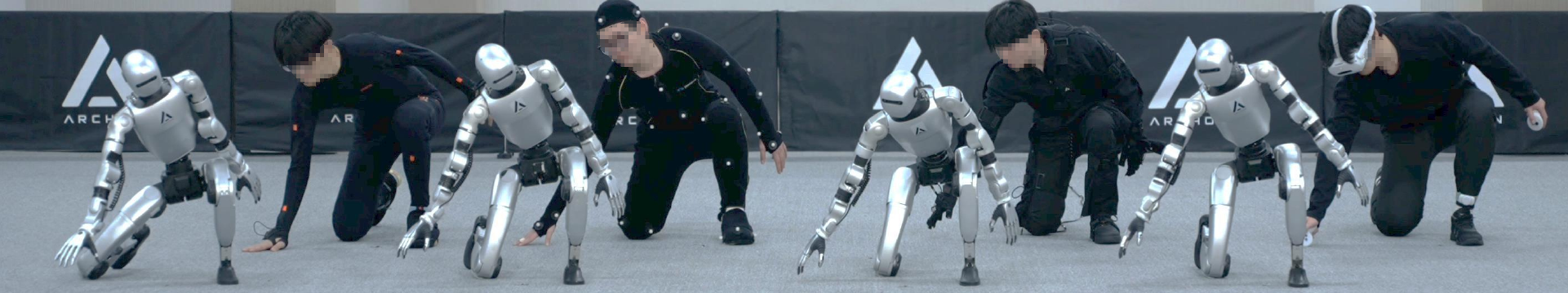}
    \caption{\textbf{Precise and stable teleoperation across operators and input devices.}
    \textbf{(A)} Comparison with the baseline during lateral reaching toward the ankle, small-amplitude lateral translation, ankle rotation, and bending forward to reach the ground.
    \textbf{(B)} Adaptation to operators with heights of 165--188~cm and body masses of 50--80~kg while maintaining stable control and preserving human motion intent.
    Labels show height (cm) / body mass (kg).
    \textbf{(C)} Support for different motion-capture devices by computing $H_5'$ directly, without retargeting.
    From left to right: Xsens (inertial motion capture), CHINGMU (optical motion capture), Noitom Perception Neuron (PN) Link (wired inertial motion capture), and PICO (virtual reality, VR).}
    \label{fig:teleop-examples}
\end{figure}

Effective teleoperation requires a whole-body controller to track the human reference accurately while maintaining stable robot behavior.
Imprecise tracking forces the operator to make repeated corrective movements, which can reduce collection efficiency, task success, and demonstration trajectory quality.
\cref{fig:teleop-examples}A qualitatively compares \modelname with the baseline across four motions that probe tracking precision, stability, and preservation of human intent: small-amplitude lateral translation, ankle rotation, bending forward to reach the ground, and lateral reaching toward the ankle.
During small-amplitude movements, \modelname produces the corresponding robot motion, whereas the baseline does not clearly reproduce these subtle changes.
During standing ankle rotation, \modelname lifts the foot as intended, whereas the baseline does not, highlighting the importance of precise end-effector tracking.
When the operator bends forward to reach the ground, \modelname exhibits few unintended steps, while the baseline begins to sway and lose stability.
During the lateral reach toward the same-side leg, the posture produced by \modelname closely matches the human reference, whereas the baseline deviates, illustrating better preservation of the intended whole-body motion.

The examples in \cref{fig:teleop-examples}B further show that \modelname accommodates operators with heights ranging from 165 to 188~cm and body weight ranging from 50 to 80~kg while maintaining stable behavior and preserving their motion intent.
We demonstrate compatibility with four motion-capture input systems: Xsens inertial motion capture, CHINGMU optical motion capture, Noitom Perception Neuron Link wired inertial motion capture (PN-LINK), and PICO virtual reality (VR) tracking (\cref{fig:teleop-examples}C).
Although these sources differ in skeleton representation and motion-capture quality, their measurements are converted into $H_5'$ and supplied directly to the policy without human-to-robot retargeting.
Across the illustrated input domains, \modelname exhibits stable and precise behavior during large-amplitude motions.

\begin{figure}[!t]
    \centering
    \includegraphics[width=\linewidth,height=0.70\textheight,keepaspectratio]{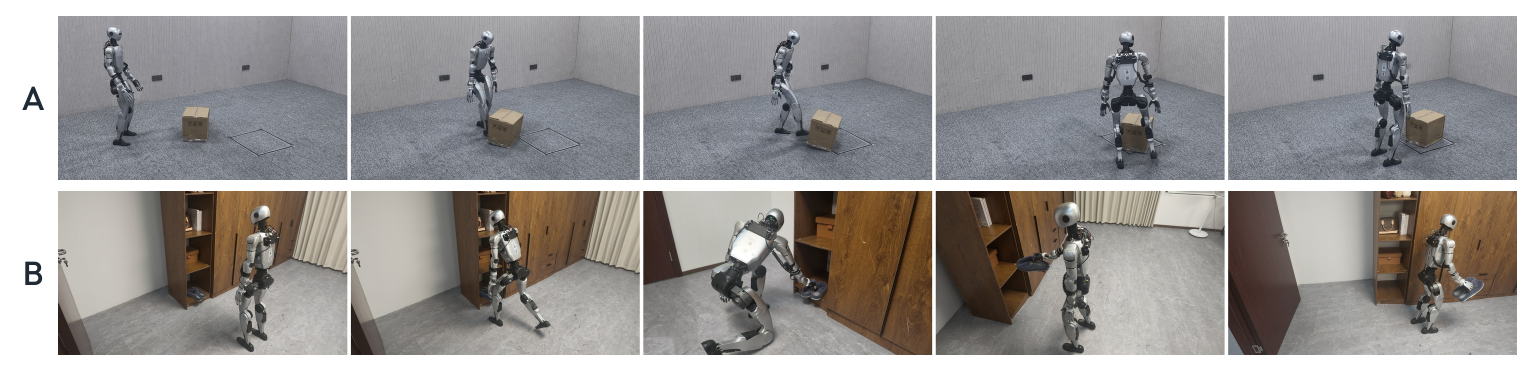}
    \caption{\textbf{Autonomous execution of downstream whole-body manipulation tasks.} The robot autonomously performs two tasks: \textbf{(A)} kicking a cardboard box into a marked target area and \textbf{(B)} retrieving slippers from a low shelf and placing them at a designated location. Each row shows the progression from approach through object interaction to task completion, with time advancing from left to right.}
    \label{fig:posttrain}
\end{figure}

To quantify demonstration collection performance, we evaluate two tasks: kicking a cardboard box into a marked target area and retrieving slippers from a low shelf before carrying and placing them at a designated location  (\cref{fig:posttrain}).
These tasks require accurate foot placement and low-reaching grasping, respectively, together with coordinated locomotion and balance across changes in whole-body posture.
\modelname yields 43 successful demonstrations per hour for kicking a cardboard box and 62 for picking up slippers, compared with 9 and 41 for SONIC v1.1 (SMPL), respectively (\cref{tab:collection-current}).
Collection success rates also increase from 37.50\% to 98.03\% for kicking and from 68.33\% to 97.37\% for slipper retrieval.

\begin{table}[!t]
\centering
\caption{\textbf{Teleoperation data collection efficiency.}
Successful demonstrations collected per hour and collection success rates for each task.
\modelname\unskip$^{*}$ denotes the variant with error feedback supplied to the encoder.
Higher values are better.}
\label{tab:collection-current}
\small
\setlength{\tabcolsep}{5pt}
\begin{tabular}{llrr}
\toprule
Controller & Task & Successful demos/h $\uparrow$ & Collection SR (\%) $\uparrow$ \\
\midrule
SONIC v1.1 (SMPL) & Kick cardboard box & 9 & 37.50 \\
 & Pick up slippers & 41 & 68.33 \\
\midrule
\modelname\unskip$^{*}$ & Kick cardboard box & 45 & 94.74 \\
 & Pick up slippers & 67 & 93.10 \\
\midrule
\modelname & Kick cardboard box & 43 & 98.03 \\
 & Pick up slippers & 62 & 97.37 \\
\bottomrule
\end{tabular}

\end{table}

\begin{table}[!t]
\centering
\caption{\textbf{Downstream task performance by inference controller and policy interface.} SR denotes success rate. Score is weighted by the number of kicks for Kick box and averaged over evaluation steps for Pick up slippers. Both metrics range from 0 to 1. \modelname\unskip$^{*}$ uses encoder error feedback. Best results are bolded, including ties.}
\label{tab:posttrain}
\label{tab:posttrain-current}
\small
\setlength{\tabcolsep}{4pt}
\renewcommand{\arraystretch}{1.0}
\begin{tabular}{@{}llcccc@{}}
\toprule
Controller (mode) & Interface & \multicolumn{2}{c}{Kick box} & \multicolumn{2}{c}{Pick up slippers} \\
\cmidrule(lr){3-4}\cmidrule(lr){5-6}
& & SR & Score & SR & Score \\
\midrule
SONIC v1.1 (SMPL) & Motion token & $0.70 \pm 0.14$ & $0.63 \pm 0.08$ & $0.10 \pm 0.00$ & $0.44 \pm 0.05$ \\
\midrule
\modelname\unskip$^{*}$ (re-locate) & Sparse & $0.85 \pm 0.07$ & $0.79 \pm 0.05$ & $\mathbf{0.45 \pm 0.07}$ & $\mathbf{0.74 \pm 0.05}$ \\
\modelname\unskip$^{*}$ (re-locate) & Motion token & $0.00 \pm 0.00$ & $0.00 \pm 0.00$ & $0.00 \pm 0.00$ & $0.00 \pm 0.00$ \\
\midrule
\modelname (re-locate + Recon. feedback) & Sparse & $0.75 \pm 0.07$ & $0.56 \pm 0.03$ & $0.20 \pm 0.14$ & $0.68 \pm 0.04$ \\
\modelname (re-locate + Recon. feedback) & Motion token & $\mathbf{0.90 \pm 0.00}$ & $\mathbf{0.81 \pm 0.02}$ & $\mathbf{0.45 \pm 0.07}$ & $0.64 \pm 0.05$ \\
\bottomrule
\end{tabular}
\end{table}

\subsubsection{Autonomous task execution with sparse and motion-token interfaces}
\label{sec2:subsec:teleop-vla:subsubsec:autonomy}
We next evaluate vision-based autonomous execution using policies trained on teleoperated data (\cref{fig:posttrain}).

For high-level policy, we fine-tune GR00T N1.7~\cite{nvidia2025gr00tn1openfoundation} on teleoperated demonstrations using flow matching, conditioned on language instructions, three RGB views, and proprioception.
The sparse-interface policy predicts 50-step action chunks containing five target poses and 12-dimensional hand commands, totaling 57 dimensions per step with 6D rotation representations.
Pelvis trajectories use a fixed chunk-anchor heading frame, while limb poses use each target frame's pelvis heading.
At deployment, these predictions form rolling four-frame reference windows dispatched at 50~Hz with 60~ms look-ahead.
The frozen tracker converts the references into whole-body joint commands, while hand commands are transmitted separately.
Real-time chunking accommodates asynchronous policy inference.

The comparison is made between SONIC v1.1 using motion tokens and \modelname in re-locate mode using either sparse references or motion tokens.
For \modelname, we evaluate encoder error feedback, denoted by \modelname\unskip$^{*}$, and feedback computed from reconstructed near-term targets, denoted by Recon. feedback.
The latter separates motion encoding from execution-dependent tracking errors: the VLA can predict motion tokens directly, bypassing the encoder, while reconstructed targets provide tracking-error feedback to the shared control decoder.
This enables motion-token execution without accompanying sparse references.

Success rate and a task-specific Score in \Cref{tab:posttrain} are reported over 20 trials.
With sparse references, \modelname\unskip$^{*}$ substantially improves autonomous task performance over the SONIC v1.1 motion-token baseline, highlighting the effectiveness of \modelname control for downstream manipulation.
Recon. feedback further enables motion tokens as a direct policy interface while achieving success rates comparable to the encoder-feedback sparse configuration.
In contrast, directly commanding motion tokens with the encoder-feedback variant yields no successful trials on either task.
These results show that Recon. feedback extends the controller to a learned motion-token interface without requiring accompanying sparse references or compromising the observed task success of sparse-reference control.

\FloatBarrier

\section{Discussion}
\label{sec3:discussion}

\subsection{Implications for humanoid whole-body manipulation}
\label{sec3:subsec:implications}
Precise and stable whole-body manipulation requires commands that preserve the intended motion and controllers that maintain accuracy as posture and support change. The reference comparisons and hardware experiments show how errors at either stage can disrupt execution. Sparse commands specify task-relevant body poses while leaving the remaining joints free to coordinate. This freedom makes the training distribution consequential: the same commanded poses can admit different whole-body configurations. 
In Anchor 1, calibrated sparse references preserve task-relevant human motion cues, while curated robot references guide whole-body coordination during training. This division retains a sparse deployment interface without requiring a complete robot joint trajectory.

The hardware comparisons show that conclusions about controller reliability depend on the precision demanded by the evaluation. Tighter position tolerances reveal differences that permissive success criteria can obscure. 
Mean tracking error summarizes average accuracy but can obscure brief deviations; SR reports the fraction of motions that satisfy the prescribed criteria throughout the evaluated window.
The three-track benchmark translates manipulation demands into explicit tests of precision and stability. Its task structure and prescribed spatial targets help identify the coordination conditions under which accuracy can be sustained, while the Public track tests a broader range of motion behaviors.

Feedback placement has consequences beyond tracking. Supplying explicit execution errors directly to the decoder, with feedback targets provided by sparse references or reconstructed from motion tokens, allows the same controller to correct deviations under both command interfaces. 
In the autonomous tasks, token commands with reconstructed-target feedback support successful execution, whereas the tested encoder-feedback variant fails when commanded through tokens. These results support separating motion commands from the explicit error signals used to execute them. These VLA-driven tasks demonstrate the use of this interface for object interaction involving locomotion and changes in posture. Precise and stable whole-body control can thus provide a common execution layer for human motion input and learned task commands.

\subsection{Limitations}
\label{sec3:subsec:limitations}
The present evaluation covers a single humanoid embodiment, leaving transfer across robot morphologies untested.
Sparse kinematic targets do not explicitly describe object geometry, contact forces, or grasp stability, so accurate motion tracking alone does not ensure successful object interaction.
The reconstruction branch also remains a source of tracking error, and further improvements are needed to narrow the precision gap between reconstructed and supplied feedback targets.

\subsection{Future directions}
\label{sec3:subsec:future}
Future work should focus on connecting whole-body tracking more directly to object interaction through object-relative targets and contact-aware control and evaluation. Testing across robot embodiments, payloads, and contact conditions would help establish the limits of transfer. For motion-token control, further work should examine how target-reconstruction errors and high-level prediction errors affect closed-loop execution, and how their impact on precision can be reduced, particularly during longer tasks with changing contacts.
\Needspace{9\baselineskip}
\section{Materials and methods}
\label{sec4:method}

\subsection{Overview of distribution curation and control}
\label{sec4:subsec:overview}
\begin{figure}[t]
    \centering
    \includegraphics[width=\linewidth]{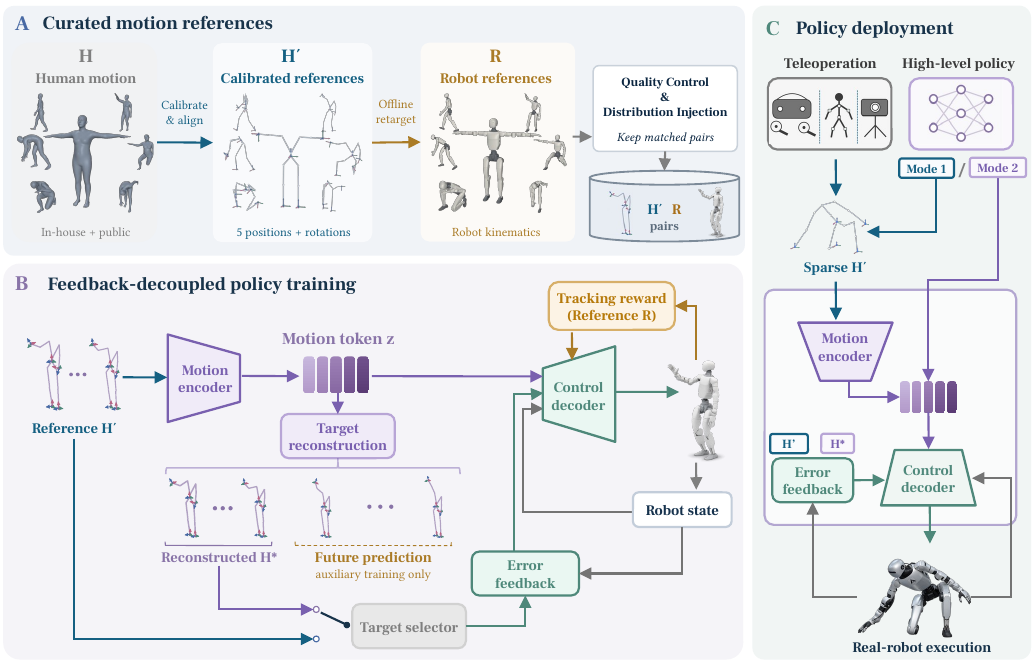}
    \caption{\textbf{Overview of \modelname.}
    \textbf{(A)} Human motion $H$ is calibrated to obtain $H'$ and retargeted offline to obtain robot references $R$; quality control retains matched pairs and shapes the training distribution.
    \textbf{(B)} The motion encoder produces tokens for control and target reconstruction.
    A target selector chooses supplied or reconstructed poses for explicit error feedback, which enters the control decoder alongside robot state.
    Future prediction is used for training only.
    \textbf{(C)} Teleoperation supplies sparse motion references; higher-level policies can supply sparse references (Mode 1) or motion tokens (Mode 2).
    The corresponding feedback targets are supplied references ($H'$) or reconstructed poses ($H^*$).}
    \label{fig:overview}
\end{figure}

\modelname combines curated motion references with a controller that separates motion encoding from explicit tracking-error feedback (\cref{fig:overview}).
Human motion is geometrically calibrated to form $H'$, from which five target poses $H'_5$ are extracted, and paired with an offline-retargeted robot reference $R$.
Quality screening retains corresponding pairs, shaping the robot behaviors available for learning while preserving sparse human references as the command interface.

The controller encodes the sparse references into motion tokens and combines these tokens with proprioception and tracking errors to generate joint actions.
A reconstruction head supplies feedback targets when explicit pose references are unavailable, enabling a higher-level policy to command the same controller through motion tokens.

\begingroup
\renewcommand{\topfraction}{0.9}
\renewcommand{\bottomfraction}{0.8}
\renewcommand{\textfraction}{0.1}
\subsection{Training data and quality control}
\label{sec4:subsec:training-data}
\label{sec4:subsec:training-data:contribution}

\subsubsection{Training dataset}
\label{sec4:subsec:training-data:subsubsec:dataset}
The training dataset combines in-house motion-capture data with motions from public datasets (\Cref{tab:training-composition}). 
After augmentation with left--right mirrored motions, it contains 406,222 sequences totaling approximately 116.8 million frames and 1,081.4 hours.
In-house and public data contribute 835.7 hours (77.3\%) and 245.7 hours (22.7\%), respectively.
Frame counts reflect the stored sampling rates, and durations are calculated using each sequence's frame rate.

Chingmu and Nokov recordings are collected using the tag-based task composition procedure described in \cref{sec4:subsubsec:inhouse}, whereas PN-LINK (locomotion) recordings were collected for tasks intended for VLA learning using a separate protocol.

We use left--right mirroring to expand the training dataset and mitigate potential bias associated with handedness and preferred movement directions.
Mirroring contributes 142,342 retained sequences, comprising approximately 42.4 million frames and 393.0 hours.
Original and mirrored sequences are counted separately.

\begin{table}[t]
\centering
\caption{\textbf{Composition of the training dataset.}
Sequence counts include original and mirrored sequences.
Duration percentages are calculated relative to the total duration after mirroring.
Totals are computed before rounding.}
\label{tab:training-composition}
\footnotesize
\setlength{\tabcolsep}{4pt}
\renewcommand{\arraystretch}{1.08}
\begin{tabular}{@{}lrrrrr@{}}
\toprule
Source & Sequences & \shortstack{Original\\duration (h)} & \shortstack{Mirrored\\duration (h)} & \shortstack{Total\\duration (h)} & \shortstack{Duration\\share (\%)} \\
\midrule
\textbf{In-house} & \textbf{288,215} & \textbf{496.04} & \textbf{339.62} & \textbf{835.67} & \textbf{77.28} \\
\quad Chingmu & 222,400 & 227.97 & 217.19 & 445.16 & 41.17 \\
\quad Nokov & 7,149 & 13.67 & 15.61 & 29.27 & 2.71 \\
\quad PN-LINK (locomotion) & 58,666 & 254.41 & 106.83 & 361.24 & 33.41 \\
\addlinespace[3pt]
\textbf{Public} & \textbf{118,007} & \textbf{192.33} & \textbf{53.37} & \textbf{245.70} & \textbf{22.72} \\
\quad BONES-SEED & 74,371 & 138.29 & 0.00 & 138.29 & 12.79 \\
\quad CIRCLE & 10,497 & 6.84 & 6.74 & 13.57 & 1.26 \\
\quad OMOMO & 7,859 & 6.51 & 6.43 & 12.94 & 1.20 \\
\quad AMASS & 25,126 & 36.09 & 35.61 & 71.70 & 6.63 \\
\quad LAFAN & 154 & 4.60 & 4.60 & 9.20 & 0.85 \\
\addlinespace[3pt]
\textbf{Total} & \textbf{406,222} & \textbf{688.37} & \textbf{392.99} & \textbf{1,081.37} & \textbf{100.00} \\
\bottomrule
\end{tabular}
\end{table}

\subsubsection{Quality control}
\label{sec4:subsec:training-data:subsubsec:qc}
All training motion sequences undergo automated quality control during data conversion, with original and left--right mirrored sequences screened individually.
Checks on retargeted robot motions assess prolonged floating, joint discontinuities, self-collision, periodic joint jitter, and ground penetration using kinematics and collision information from the robot model.
Temporal criteria account for each sequence's sampling rate.

All in-house sequences additionally undergo human review of retargeted motion playback, with source human-skeleton animations consulted to assess motion quality and correspondence.
In-house sequences are retained only if they pass both automated checks and human review; public sequences are retained if they pass automated checks.

\Needspace{7\baselineskip}
\subsection{Benchmark construction}
\label{sec4:subsec:benchmark}

\subsubsection{In-house track}
\label{sec4:subsubsec:inhouse}
\begin{table}[t]
\centering
\caption{\textbf{First-level and second-level reference tags for coverage analysis.}
U and L denote upper-body and lower-body descriptors, respectively; 1 and 2 indicate levels of description.}
\label{tab:inhouse-action-tags}
\footnotesize
\setlength{\tabcolsep}{4pt}
\renewcommand{\arraystretch}{1.08}
\begin{minipage}[t]{0.485\linewidth}
\vspace{0pt}
\setlength{\tabcolsep}{3pt}
\begin{tabular}{@{}p{0.17\linewidth}p{\dimexpr0.83\linewidth-2\tabcolsep\relax}@{}}
\toprule
\multicolumn{2}{@{}l@{}}{\textbf{Upper body}} \\
\midrule
\multicolumn{2}{@{}l@{}}{\textit{U1: Torso configuration}} \\
\addlinespace[2pt]
\texttt{U1-1} & \raggedright Upright torso \tabularnewline
\texttt{U1-2} & \raggedright Forward bending \tabularnewline
\texttt{U1-3} & \raggedright Lateral bending \tabularnewline
\texttt{U1-4} & \raggedright Torso rotation \tabularnewline
\texttt{U1-5} & \raggedright Forward bending with rotation \tabularnewline
\texttt{U1-6} & \raggedright Lateral bending with rotation \tabularnewline
\texttt{U1-7} & \raggedright Combined forward and lateral bending \tabularnewline
\texttt{U1-8} & \raggedright Torso coordination during sit--stand transitions \tabularnewline
\addlinespace[5pt]
\multicolumn{2}{@{}l@{}}{\textit{U2: Arm and hand behavior}} \\
\addlinespace[2pt]
\texttt{U2-1} & \raggedright Arms resting at the sides \tabularnewline
\texttt{U2-2} & \raggedright Unilateral reaching \tabularnewline
\texttt{U2-3} & \raggedright Symmetric bimanual reaching \tabularnewline
\texttt{U2-4} & \raggedright Asymmetric bimanual reaching \tabularnewline
\texttt{U2-5} & \raggedright Elbow flexion \tabularnewline
\texttt{U2-6} & \raggedright Unilateral arm swinging / sweeping \tabularnewline
\texttt{U2-7} & \raggedright Same-direction bilateral arm swinging \tabularnewline
\texttt{U2-8} & \raggedright Opposite-direction bilateral arm swinging \tabularnewline
\texttt{U2-9} & \raggedright Alternating arm swinging \tabularnewline
\texttt{U2-10} & \raggedright World-fixed single-hand hold \tabularnewline
\bottomrule
\end{tabular}
\end{minipage}
\hfill
\begin{minipage}[t]{0.485\linewidth}
\vspace{0pt}
\setlength{\tabcolsep}{3pt}
\begin{tabular}{@{}p{0.17\linewidth}p{\dimexpr0.83\linewidth-2\tabcolsep\relax}@{}}
\toprule
\multicolumn{2}{@{}l@{}}{\textbf{Lower body}} \\
\midrule
\multicolumn{2}{@{}l@{}}{\textit{L1: Support and movement modes}} \\
\addlinespace[2pt]
\texttt{L1-1} & \raggedright Stationary standing \tabularnewline
\texttt{L1-2} & \raggedright Omnidirectional locomotion \tabularnewline
\texttt{L1-3} & \raggedright Squatting \tabularnewline
\texttt{L1-4} & \raggedright Half kneeling \tabularnewline
\texttt{L1-5} & \raggedright Lower-body manipulation \tabularnewline
\texttt{L1-6} & \raggedright Sit--stand transitions \tabularnewline
\addlinespace[5pt]
\multicolumn{2}{@{}l@{}}{\textit{L2: Specific stances and movement patterns}} \\
\addlinespace[2pt]
\texttt{L2-1} & \raggedright Side-by-side stance \tabularnewline
\texttt{L2-2} & \raggedright Staggered stance \tabularnewline
\texttt{L2-3} & \raggedright Turning \tabularnewline
\texttt{L2-4} & \raggedright Small forward steps \tabularnewline
\texttt{L2-5} & \raggedright Small lateral steps \tabularnewline
\texttt{L2-6} & \raggedright Straight-line forward walking \tabularnewline
\texttt{L2-7} & \raggedright Lateral walking \tabularnewline
\texttt{L2-8} & \raggedright Backward walking \tabularnewline
\texttt{L2-9} & \raggedright Diagonal walking \tabularnewline
\texttt{L2-10} & \raggedright Circular walking \tabularnewline
\texttt{L2-11} & \raggedright Shallow squat \tabularnewline
\texttt{L2-12} & \raggedright Deep squat \tabularnewline
\texttt{L2-13} & \raggedright Foot reaching \tabularnewline
\texttt{L2-14} & \raggedright Stepping over obstacles \tabularnewline
\texttt{L2-15} & \raggedright Side-on passage through narrow spaces\tabularnewline
\texttt{L2-16} & \raggedright Lateral stepping \tabularnewline
\texttt{L2-17} & \raggedright Forward stepping \tabularnewline
\texttt{L2-18} & \raggedright Floor sit--stand transitions \tabularnewline
\texttt{L2-19} & \raggedright Chair sit--stand transitions \tabularnewline
\bottomrule
\end{tabular}
\end{minipage}
\end{table}

We construct the In-house track from in-house motion-capture tasks with separate descriptions of upper- and lower-body actions.
U1 and U2 describe torso configuration and arm-and-hand behavior, respectively.
L1 and L2 describe support and movement modes and specific stance and movement patterns, respectively.
This separation makes the intended hand behavior and its whole-body context explicit, allowing similar manipulation demands to be specified under different postural and locomotor conditions.

The first-level and second-level tags describe complementary action dimensions. Meanwhile, we note that they do not form a strict parent--child taxonomy.
Their combinations follow predefined compatibility relations, allowing a second-level tag to occur with multiple compatible first-level tags.
The vocabulary covers upright, bent, and rotated torso configurations; unilateral and bilateral reaching, arm swinging, and end-effector holding; and stationary support, stepping, squatting, and kneeling.

Coverage analysis uses a reference set of 43 first-level and second-level tags derived from previously defined valid motion-capture tasks (\cref{tab:inhouse-action-tags}).
Sit--stand descriptors remain in this reference set despite being absent from the current In-house track.

\begin{figure}[t]
    \centering
    \includegraphics[width=\linewidth]{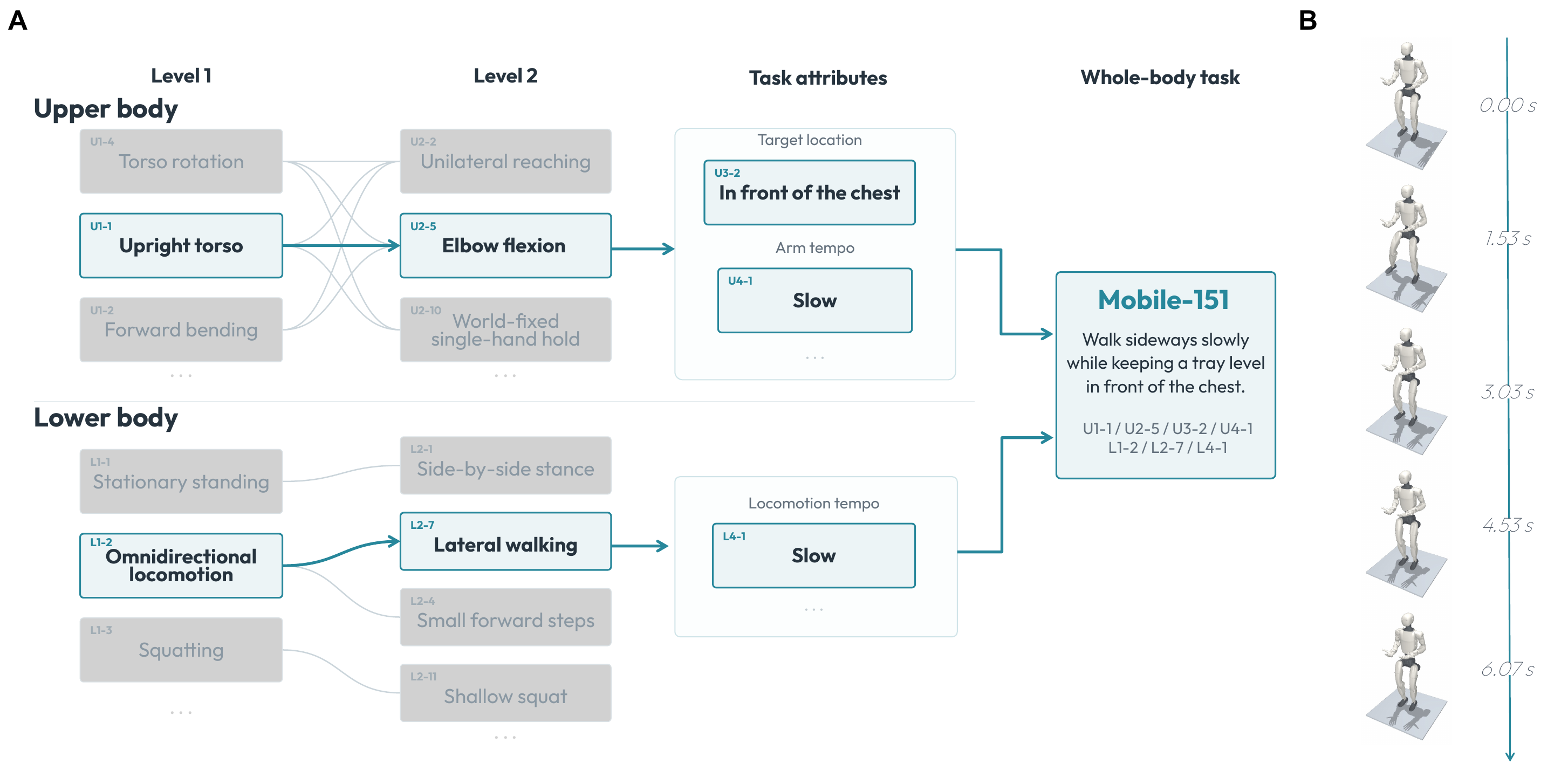}
    \caption{\textbf{Task construction and reference motion for Mobile-151.}
    \textbf{(A)} upper-body and lower-body action descriptions and task attributes define the task.
    \textbf{(B)} Representative reference poses, with timestamps in seconds.}
    \label{fig:inhouse-task-construction}
\end{figure}

Task construction follows predefined compatibility relations that determine admissible refinements within each action description and allowable combinations of upper-body and lower-body actions.
Combinations that violate these relations are excluded, yielding a constrained task library rather than the unrestricted Cartesian product of the tag sets.
Task attributes specify target locations, movement events, and tempo where applicable.
Third- and fourth-level descriptors are included according to the task requirements; they do not form a mandatory sequential refinement chain and thus are omitted.

The selected action descriptions and attributes are translated into motion-capture collection instructions.
\cref{fig:inhouse-task-construction} illustrates this process for Mobile-151.
The upper-body description specifies an upright torso, flexed elbows, a hand position in front of the chest, and a slow arm tempo; the lower-body description specifies lateral walking at a slow tempo.
The resulting instruction is to walk sideways slowly while keeping a tray level in front of the chest.

\begin{table}[t]
\centering
\caption{\textbf{Sources and selection protocols for the Public track.}}
\label{tab:public-benchmark}
\footnotesize
\setlength{\tabcolsep}{5pt}
\renewcommand{\arraystretch}{1.08}
\begin{tabular}{@{}llr@{}}
\toprule
Source & Selection protocol & Sequences \\
\midrule
BONES-SEED~\cite{bones_seed_dataset} & Held-out set of 18 actors & 1,214 \\
CIRCLE~\cite{araujo2023circle} & Held-out subject s5 & 442 \\
OMOMO~\cite{li2023object} & Existing test partition & 474 \\
AMASS~\cite{mahmood2019amass} & Predefined local test subset & 100 \\
\addlinespace[2pt]
\textbf{Total} & & \textbf{2,230} \\
\bottomrule
\end{tabular}
\end{table}

\subsubsection{Public track}
\label{sec4:subsec:benchmark:subsubsec:public}
We assemble the Public track from the public datasets BONES-SEED~\cite{bones_seed_dataset}, CIRCLE~\cite{araujo2023circle}, OMOMO~\cite{li2023object}, AMASS~\cite{mahmood2019amass} and LAFAN~\cite{harvey2020robust} using source-specific selection protocols, as depicted in \Cref{tab:public-benchmark}.
For identity-based selection, we retain all available sequences from the designated actors or subject in the dataset version used.
The resulting track contains 2,230 motion sequences totaling 3.59 hours.

Selected sequences are removed from the training pool.
Their mirrored counterparts from CIRCLE~\cite{araujo2023circle}, OMOMO~\cite{li2023object}, and AMASS~\cite{mahmood2019amass} are also excluded from training.

\subsubsection{Reachability track}
\label{sec4:subsec:benchmark:subsubsec:reachability}

We construct the Reachability track from fixed Cartesian wrist targets and corresponding whole-body reference motions across five postures: standing, shallow squat, deep squat, staggered stance, and half kneeling.
Targets are specified in meters in a world coordinate frame with $x$ forward, $y$ leftward, and $z$ upward.
We first generate right-wrist references on a coarse 20-cm Cartesian grid and add 10-cm samples to extend spatial coverage.
Each case pairs a target with a posture, so the same spatial target can occur in multiple posture groups.

We first generate base motions for each posture using Kimodo~\cite{rempe2026kimodo}, then adapt these motions to the wrist targets using constrained inverse kinematics (IK), subject to joint limits and prescribed support contacts.
The six wrist roll, pitch, and yaw joint angles are fixed at zero in the reference.
Each reference contains 464 samples at 50~Hz and spans 0--9.26~s.
The sequence includes posture preparation, a reach from 2.75 to 5~s, target holding from 5 to 7~s, and a return from 7 to 9.26~s.
We require the maximum right-wrist position error relative to the target to remain below 1~cm over $[5,7)$~s.
Tracking and foot-event evaluation use the broader task window $[2.76,9.26]$~s (\cref{sec4:subsec:evaluation:subsubsec:metrics}).

All generated reference motions additionally undergo manual quality control (\cref{sec4:subsec:training-data:subsubsec:qc}).
We mirror the accepted references across the sagittal plane, exchanging left and right limbs and reflecting the targets to include reaching with either wrist.
The final track contains 3,210 references, comprising 1,605 original motions and 1,605 mirrored counterparts.

\FloatBarrier
\endgroup
\Needspace{7\baselineskip}
\subsection{Evaluation metrics and simulation settings}
\label{sec4:subsec:evaluation}

\subsubsection{Metric definitions and aggregation}
\label{sec4:subsec:evaluation:subsubsec:metrics}

For the simulation tracks in \cref{sec2:subsec:sim}, we report success rate (SR), mean position errors, and reachability foot-event counts.
The global and local scoring rules, indexed by $f\in\{\mathrm g,\mathrm l\}$, are independent of the controller's input mode.

For one motion, let $\bp_b(t)$ be the executed position after fixed initial pelvis XY/yaw alignment and $\bar{\bp}_b(t)$ the corresponding reference position.
Here $r$ denotes the pelvis and $t=1,\ldots,T$ indexes control frames at $\Delta t=0.02$~s.
Both scores use the same reference trajectory; local error subtracts each trajectory's current pelvis XY position when computing the difference.
With $\Pi_{xy}=\operatorname{diag}(1,1,0)$, the errors are
\begin{equation}
\begin{aligned}
e_b^{\mathrm g}(t)
&= \left\|\bp_b(t)-\bar{\bp}_b(t)\right\|_2, \\
e_b^{\mathrm l}(t)
&= \left\|
\left[\bp_b(t)-\Pi_{xy}\bp_r(t)\right]
-\left[\bar{\bp}_b(t)-\Pi_{xy}\bar{\bp}_r(t)\right]
\right\|_2, \\
\theta_b(t)
&= \arccos\!\left(
\frac{\operatorname{tr}\!\left[
\bar{\bR}_b(t)^{\mathsf T}\bR_b(t)
\right]-1}{2}\right).
\end{aligned}
\label{eq:tracking-deviations}
\end{equation}
Here $\bR_b$ is the executed orientation after the same fixed yaw alignment and $\bar{\bR}_b$ the reference orientation; both scores use the same orientation error.
Errors compare the robot state after each control step with the reference at the same time, preserving height and heading differences.

Let $\mathcal B$ denote the tracked body points and $\mathcal E$ the two wrists and two ankles.
For a set of control frames $\mathcal U$, success requires all three threshold conditions:
\begin{equation}
\begin{aligned}
s^f(\mathcal U)
&= \mathbf{1}\!\left[
\begin{gathered}
\max_{t\in\mathcal U,\,b\in\mathcal E}e_b^f(t)\leq0.10\,\mathrm m, \\
\max_{t\in\mathcal U,\,b\in\mathcal E}\theta_b(t)\leq1\,\mathrm{rad}, \\
\max_{t\in\mathcal U,\,b\in\mathcal B\setminus\mathcal E}e_b^f(t)\leq0.25\,\mathrm m
\end{gathered}
\right], \\
\mathrm{SR}^{f}
&= \frac{100}{N}\sum_{i=1}^{N}s_i^f(\{1,\ldots,T_i\})\quad(\%).
\end{aligned}
\label{eq:tracking-sr}
\end{equation}
The indicator $\mathbf{1}[\cdot]$ is one when all enclosed conditions hold and zero otherwise; subscript $i$ identifies a motion when aggregating over the dataset.
Each of the $N$ motions contributes equally, and failed motions remain in the denominator.

Mean position errors use a separate failure rule: any of the 14 tracked body points exceeding 25~cm ends the observation window, independently of the SR conditions.
This more permissive criterion reduces premature truncation under the stricter SR conditions, retaining more frames to characterize tracking errors and improving temporal coverage for comparisons across controllers.
For each motion and scoring frame, we average position errors over the selected body points and all frames preceding the first threshold violation, excluding the violating frame.
The full motion is used if no violation occurs.
These per-motion means are then averaged with equal weight across motions and reported in millimeters; motions without retained frames are excluded.
Whole-body error averages all 14 tracked points, while wrist and ankle errors average the corresponding left and right points in the local scoring frame.
Root error averages the Euclidean distance between the executed and reference pelvis positions in the horizontal plane, using the global scoring frame and its corresponding observation window.
It uses the same equal-motion aggregation as the other position errors.

For the Reachability track, we measure Slip and Extra step over the task window defined in \cref{sec4:subsec:benchmark:subsubsec:reachability}.
Slip counts sustained sliding of a supporting foot, with at most one event per continuous contact phase.
Extra step counts foot-lift events that begin within the task window while the reference requires continuous ground contact throughout the event.
Both metrics sum events from the two feet and are averaged equally across the 3,210 references.
Foot-event counts use the complete task window even when tracking fails.
Reachability task SR requires global tracking success throughout the task window under \cref{eq:tracking-sr}, with no slips or extra steps.

\subsubsection{Domain randomization}
\label{sec4:subsec:evaluation:subsubsec:simulation}

We use the common domain-randomization (DR) protocol in \Cref{tab:evaluation-dr}, pairing corresponding body and joint perturbations across methods.

\begin{table}[t]
\centering
\small
\caption{\textbf{Domain randomization and sampling settings for simulation evaluation.}
DR denotes domain randomization and CoM denotes center of mass.
$\mathcal U[a,b]$ denotes uniform sampling on $[a,b]$.
Randomized physical parameters are applied before each motion and remain fixed throughout its rollout.}
\label{tab:evaluation-dr}
\begin{tabular}{@{}p{0.33\linewidth}p{0.61\linewidth}@{}}
\toprule
Parameter & Evaluation setting \\
\midrule
Random seed & 0; deterministic sampling by motion and parameter \\
Robot static friction & $\mathcal U[0.3,1.6]$ \\
Robot dynamic friction & $\mathcal U[0.3,1.2]$ \\
Robot restitution & $\mathcal U[0,0.5]$ \\
Contact-material assignment & 64 shared material buckets, assigned per collision shape \\
Ground material & Friction 1; restitution 0; fixed \\
Body mass and inertia & Independent $\mathcal U[0.8,2.5]$ scale for the torso and each wrist; inertia scaled with mass \\
Torso CoM offset & $x:\mathcal U[-0.025,0.025]$~m; $y,z:\mathcal U[-0.05,0.05]$~m \\
Default joint-position offset & $\mathcal U[-0.01,0.01]$~rad per joint, applied to nominal motor targets \\
External pushes & Disabled \\
Observation noise & Disabled \\
\bottomrule
\end{tabular}
\end{table}

\Needspace{8\baselineskip}
\subsection{Sparse human references and decoupled error feedback}
\label{sec4:subsec:formulation}

\subsubsection{Sparse motion references and offline distribution curation}
\label{sec4:subsec:formulation:subsubsec:target}
\label{sec4:subsec:formulation:subsubsec:curation}

Human and robot skeletons differ in body proportions and local coordinate conventions.
We account for these differences by calibrating human motion against the robot's T-pose to construct a structure-preserving skeletal representation $H'$ at robot scale.
The calibration uses human and robot T-pose references to determine segment- and body-specific scale factors, rotational alignments, and positional offsets, without solving for robot joint configurations.
These parameters remain fixed for each calibrated skeleton and are applied frame by frame to obtain the positions and orientations of the corresponding links.

From $H'$, we extract the positions and orientations of the pelvis, both wrists, and both ankles to form the five-pose sparse motion references $H'_5$.
This interface is designed to focus control on end-effector precision and sustained pose tracking, which are central to many whole-body manipulation tasks, while retaining flexibility in whole-body coordination.
Leaving the remaining joint configuration unspecified preserves kinematic redundancy, allowing the policy to adjust the trunk and limbs to meet the commanded poses.
Execution therefore requires learned whole-body motion completion: the policy must infer coordinated behavior from the five-pose references.
Because these references admit multiple realizations, the behavior distribution used for training is an essential part of this design.
We strictly curate the paired $H'$ and robot references $R$ to favor stable, physically plausible motions, introducing a deliberate bias toward the whole-body behaviors preferred for whole-body manipulation.
This data selection guides how the policy uses the freedom afforded by sparse commands.

For training, each $H'$ sequence is paired with a temporally aligned robot reference $R$ obtained through offline retargeting of the same human motion~\cite{nvidia_soma_retargeter}.
The sparse references $H'_5$ provide the motion input, whereas $R$ provides robot motion targets for training supervision and reward computation.
We curate the paired sequences through automated quality checks and additional manual review of in-house data, as described in \cref{sec4:subsec:training-data:subsubsec:qc}, to favor stable and coherent whole-body behavior.
At deployment, the controller uses $H'_5$ directly without requiring online retargeting to $R$, avoiding exposure to the solver failures illustrated in \cref{fig:hprime}.
This also removes the computational overhead and latency associated with online retargeting.

\subsubsection{Motion encoding, target prediction, and error feedback}
\label{sec4:subsec:formulation:subsubsec:architecture}

Explicit tracking-error feedback helps the controller correct deviations during execution~\cite{He2024omniho,Pan2025AMS,Zeng2026ScaleBFM}.
However, the same motion reference can produce different tracking errors depending on the robot's current state.
Encoding these errors into the motion token couples the representation of the command to its ongoing execution, potentially complicating token interpolation and prediction by higher-level policies.
We therefore separate explicit tracking-error feedback from motion encoding within an encoder--quantizer--decoder architecture (\cref{fig:overview}B), building on the motion-token formulation of SONIC~\cite{luo2026sonic}.
At control time $t$, we construct the encoder input $x_t$ from the global references $H'_5$ at offsets of 0, 20, 40, and 60~ms, expressing the target poses in the current robot-root frame.
The motion encoder $E$ compresses this window, and a finite scalar quantizer $Q$ produces the motion token $z_t=Q(E(x_t))$.
Explicit wrist and ankle tracking errors are excluded from the encoder input and supplied separately to the control decoder alongside proprioception.

A trajectory head $K$ decodes $z_t$ into near-term pose reconstructions and longer-horizon motion predictions.
The near-term output reconstructs the five target poses at the four input times, whereas the future output predicts changes relative to the first target pose at longer horizons.
A target selector uses either the supplied near-term references or their reconstructions to form the control feedback.
The future predictions serve only as an auxiliary training objective and are not passed to the control decoder.

We compute current wrist and ankle poses by forward kinematics from measured joint positions.
For each near-term target, we compute the position error in the current robot-root frame and the relative orientation error between the target and measured poses to form the feedback $e_t$.
The control decoder $D$ combines this feedback with the token and proprioceptive observations $o_t$:
\begin{equation}
    a_t=D(z_t,o_t,e_t).
    \label{eq:feedback-decoupled-control}
\end{equation}
Here $a_t$ denotes the joint action and $o_t$ includes robot state history.
The token therefore reaches the decoder directly, while explicit execution-error correction enters through a separate path.
When commands are supplied as motion tokens, reconstructed targets provide the feedback needed for closed-loop execution without an accompanying pose-reference sequence.

\subsubsection{Policy training and auxiliary objectives}
\label{sec4:subsec:training}

We train the controller in simulation using proximal policy optimization (PPO)~\cite{schulman2017proximal} on the curated motion pairs.
The actor is conditioned on $H'_5$, and reference-tracking rewards are evaluated against the paired robot motion $R$.
To support feedback from both supplied references and token reconstructions, we train the shared control decoder with both sources of near-term targets.
In the reconstructed-target branch, policy gradients propagate through the tracking-error computation to the reconstruction head, allowing its outputs to be optimized for control as well as pose reconstruction.
Measured poses and supplied reference targets are treated as constants in this computation.

The motion representation is trained with near-term reconstruction, future prediction, and cycle-consistency objectives in addition to PPO:
\begin{equation}
    \mathcal L=\mathcal L_{\mathrm{PPO}}
    +\lambda_{\mathrm{near}}\mathcal L_{\mathrm{near}}
    +\lambda_{\mathrm{future}}\mathcal L_{\mathrm{future}}
    +\lambda_{\mathrm{cycle}}\mathcal L_{\mathrm{cycle}}.
    \label{eq:controller-training-objective}
\end{equation}
Near-term and future supervision comes from the calibrated sparse human references $H'_5$, rather than the robot reference $R$.
The near-term objective penalizes pose-reconstruction errors, and the future objective predicts positional and rotational changes beyond the input window.
Following~\cite{luo2026sonic}, we re-encode the reconstructed near-term sparse poses with explicit tracking-error channels masked and align their bounded continuous latent with a stop-gradient latent from the original reference window to enforce cycle consistency.
These auxiliary objectives use a clean reference encoding with shared encoder, quantizer, and trajectory-head parameters.

\subsubsection{Controller deployment}
\label{sec4:subsec:deployment}

At deployment, \modelname exposes two command interfaces that share the same trained control decoder (\cref{fig:overview}C).
The sparse-reference interface accepts five target poses, which are encoded into motion tokens.
For teleoperation, these poses are obtained from motion data from motion-capture systems or virtual reality (VR) tracking devices through skeletal calibration and five-pose extraction.
Higher-level policies can use either interface, generating five-pose sparse motion references for the encoder or supplying motion tokens directly to bypass it.
In the latter case, the trajectory head reconstructs the near-term targets needed to compute tracking errors.
In both cases, the decoder combines motion tokens with proprioception and explicit tracking-error feedback to generate joint actions, without requiring the paired robot training reference $R$.

During training, the reference motion retains its global trajectory.
At inference time, global mode preserves this trajectory after an initial position and heading alignment and requires robot-root localization.
Alternatively, we apply re-locate, which treats the robot's current root position as having reached the previous reference frame's root position.
We align these two positions by translating the entire reference window:
\begin{equation}
    \widetilde{\bp}_i(t+\tau)
    =\bp_i^{\mathrm{ref}}(t+\tau)
    +\left[\bp_{\mathrm{root}}^{\mathrm{robot}}(t)
    -\bp_{\mathrm{root}}^{\mathrm{ref}}(t-\Delta t)\right].
    \label{eq:relocate-reference}
\end{equation}
Here $i$ indexes the five target bodies, $t$ is the current reference time, and $\tau\in\{0,\Delta t,2\Delta t,3\Delta t\}$ indexes the time offsets in the reference window, with $\Delta t=20$~ms.
The root denotes the pelvis, and the first reference frame serves as the anchor at the start of a sequence.
The same translation is applied to all five targets and all times in the window, preserving their relative geometry and the reference displacement from the previous frame to the current frame.
Target orientations remain unchanged.
This alignment is applied before motion encoding and is separate from the T-pose calibration of skeletal dimensions.
Both inference modes use the same trained controller without additional optimization.

\Needspace{24\baselineskip}
\section*{Acknowledgements}

We thank the data collection team at Archon Robotics for dedicated data collection, quality checks, and help with the teleoperation tests; the hardware team for hardware development and repairs; the infrastructure team for data platform and robot infrastructure support; and the operations team for project management.

\section*{Contributions}

\noindent \textbf{Core contributors:} Yixuan Pan, Caojun Wang, Penglin Fu, Guangjun Zeng, Ziyan Li

\noindent \textbf{Model and experiments:} Yixuan Pan, Penglin Fu, Guangjun Zeng, Caojun Wang, Ziyan Li, Yu Zhang
    
\noindent \textbf{Data construction and quality control:} Caojun Wang, Guangjun Zeng,
Shijia Peng

\noindent \textbf{Benchmark design:} Caojun Wang, Penglin Fu, Ziyan Li, Yixuan Pan

\noindent \textbf{Vision-language-action (VLA) integration:} Modi Shi, Jialong Zeng, Checheng Yu, Lirui Zhao

\noindent \textbf{Robot infrastructure:} Jialong Zeng, Haoran Jiang, Yu Zhang, Wencong Zhang

\noindent \textbf{Software platform and tooling:} Huibin Yan, Liang Chen, Yong Zhang

\noindent \textbf{Hardware:} Kai Cheng, Shijia Peng

\noindent \textbf{Data acquisition and logistics:} Jin Yang, Xuechong Lv, Gangcheng Jiang, Huijie Wang

\noindent \textbf{Project leads:} Li Chen (first point of contact), Tianyu Li,
Yu Qiao, Hongyang Li

\bibliographystyle{unsrtnat}
\bibliography{bibliography_long, references}

\begin{thebibliography}{41}
\providecommand{\natexlab}[1]{#1}
\providecommand{\url}[1]{\texttt{#1}}
\expandafter\ifx\csname urlstyle\endcsname\relax
  \providecommand{\doi}[1]{doi: #1}\else
  \providecommand{\doi}{doi: \begingroup \urlstyle{rm}\Url}\fi

\bibitem[Gu et~al.(2026)Gu, Li, Shen, Yu, Xie, McCrory, Cheng, Shamsah, Griffin, Liu, Kheddar, Peng, Zhu, Shi, Nguyen, Cheng, Gao, and Zhao]{gu2025humanoid}
Zhaoyuan Gu, Junheng Li, Wenlan Shen, Wenhao Yu, Zhaoming Xie, Stephen McCrory, Xianyi Cheng, Abdulaziz Shamsah, Robert Griffin, C.~Karen Liu, Abderrahmane Kheddar, Xue~Bin Peng, Yuke Zhu, Guanya Shi, Quan Nguyen, Gordon Cheng, Huijun Gao, and Ye~Zhao.
\newblock Humanoid locomotion and manipulation: Current progress and challenges in control, planning, and learning.
\newblock \emph{IEEE/ASME Trans. on Mechatronics (TMECH)}, 2026.

\bibitem[He et~al.(2024)He, Luo, He, Xiao, Zhang, Zhang, Kitani, Liu, and Shi]{He2024omniho}
Tairan He, Zhengyi Luo, Xialin He, Wenli Xiao, Chong Zhang, Weinan Zhang, Kris Kitani, Changliu Liu, and Guanya Shi.
\newblock {OmniH2O}: Universal and dexterous human-to-humanoid whole-body teleoperation and learning.
\newblock In \emph{Proc. Conf. on Robot Learning (CoRL)}, 2024.

\bibitem[Fu et~al.(2024)Fu, Zhao, Wu, Wetzstein, and Finn]{Fu2024humanplus}
Zipeng Fu, Qingqing Zhao, Qi~Wu, Gordon Wetzstein, and Chelsea Finn.
\newblock {HumanPlus}: Humanoid shadowing and imitation from humans.
\newblock In \emph{Proc. Conf. on Robot Learning (CoRL)}, 2024.

\bibitem[Ze et~al.(2026)Ze, Zhao, Wang, Kanazawa, Duan, Abbeel, Shi, Wu, and Liu]{Ze2025TWIST2}
Yanjie Ze, Siheng Zhao, Weizhuo Wang, Angjoo Kanazawa, Rocky Duan, Pieter Abbeel, Guanya Shi, Jiajun Wu, and C.~Karen Liu.
\newblock {TWIST2}: Scalable, portable, and holistic humanoid data collection system.
\newblock In \emph{Proc. IEEE International Conf. on Robotics and Automation (ICRA)}, 2026.

\bibitem[Jiang et~al.(2025)Jiang, Chen, Bu, et~al.]{jiang2025wholebodyvla}
Haoran Jiang, Jin Chen, Qingwen Bu, et~al.
\newblock Wholebodyvla: Towards unified latent vla for whole-body loco-manipulation control.
\newblock In \emph{Proc. of the International Conf. on Learning Representations (ICLR)}, 2025.

\bibitem[Cheng et~al.(2024)Cheng, Ji, Chen, Yang, Yang, and Wang]{Cheng2024exbody}
Xuxin Cheng, Yandong Ji, Junming Chen, Ruihan Yang, Ge~Yang, and Xiaolong Wang.
\newblock Expressive whole-body control for humanoid robots.
\newblock In \emph{Proc. Robotics: Science and Systems (RSS)}, 2024.

\bibitem[Chen et~al.(2025)Chen, Ji, Cheng, Peng, Peng, and Wang]{Chen2025gmt}
Zixuan Chen, Mazeyu Ji, Xuxin Cheng, Xuanbin Peng, Xue~Bin Peng, and Xiaolong Wang.
\newblock {GMT}: General motion tracking for humanoid whole-body control.
\newblock \emph{arXiv preprint arXiv:2506.14770}, 2025.

\bibitem[Qi et~al.(2026)Qi, Chen, Wang, Lin, Lian, Zhang, Yu, Wang, and Yi]{Qi2026HumanoidGPT}
Zekun Qi, Xuchuan Chen, Jilong Wang, Chenghuai Lin, Yunrui Lian, Wenyao Zhang, Xinqiang Yu, He~Wang, and Li~Yi.
\newblock Humanoid generative pre-training for zero-shot motion tracking.
\newblock In \emph{Proc. IEEE Conf. on Computer Vision and Pattern Recognition (CVPR)}, 2026.

\bibitem[Liao et~al.(2026{\natexlab{a}})Liao, Truong, Huang, Gao, Tevet, Sreenath, and Liu]{liao2026beyondmimic}
Qiayuan Liao, Takara~E Truong, Xiaoyu Huang, Yuman Gao, Guy Tevet, Koushil Sreenath, and C~Karen Liu.
\newblock {BeyondMimic}: From motion tracking to versatile humanoid control via guided diffusion.
\newblock \emph{Science Robotics}, 11\penalty0 (117), 2026{\natexlab{a}}.

\bibitem[Sleiman et~al.(2026)Sleiman, Li, Adu-Bredu, Deits, Kumar, Bergamin, Bhardwaj, Biddlestone, Burger, Estrada, Iacobelli, Koolen, Lambert, Lin, Mungai, Nobles, Rozen-Levy, Shi, Wang, Welner, Yu, Zhang, Rizzi, Hodgins, Bertrand, Abe, Kuindersma, and Farshidian]{sleiman2026zest}
Jean~Pierre Sleiman, He~Li, Alphonsus Adu-Bredu, Robin Deits, Arun Kumar, Kevin Bergamin, Mohak Bhardwaj, Scott Biddlestone, Nicola Burger, Matthew~A. Estrada, Francesco Iacobelli, Twan Koolen, Alexander Lambert, Erica Lin, M.~Eva Mungai, Zach Nobles, Shane Rozen-Levy, Yuyao Shi, Jiashun Wang, Jakob Welner, Fangzhou Yu, Mike Zhang, Alfred Rizzi, Jessica Hodgins, Sylvain Bertrand, Yeuhi Abe, Scott Kuindersma, and Farbod Farshidian.
\newblock {ZEST}: Zero-shot embodied skill transfer for athletic robot control.
\newblock \emph{Science Robotics}, 11\penalty0 (117), 2026.

\bibitem[Liao et~al.(2026{\natexlab{b}})Liao, Shao, Long, Zakka, Horowitz, Abbeel, and Sreenath]{Liao2026SuperMimic}
Qiayuan Liao, Yiyang Shao, Junfeng Long, Kevin Zakka, Roberto Horowitz, Pieter Abbeel, and Koushil Sreenath.
\newblock {SuperMimic}: Scaling humanoid whole-body control with one differentiable world model.
\newblock In \emph{Proc. Conf. on Robot Learning (CoRL)}, 2026{\natexlab{b}}.

\bibitem[Luo et~al.(2026)Luo, Yuan, Wang, Li, Castañeda, Chen, Cao, Li, Minor, Ben, Park, Sami, Wang, Da, Ding, Hogg, Song, Lim, Jeong, He, Xue, Xiao, Yuen, Kautz, Chang, Iqbal, Fan, and Zhu]{luo2026sonic}
Zhengyi Luo, Ye~Yuan, Tingwu Wang, Chenran Li, Fernando Castañeda, Sirui Chen, Zi-Ang Cao, Jiefeng Li, David Minor, Qingwei Ben, Jinhyung Park, David Sami, Zi~Wang, Xingye Da, Runyu Ding, Cyrus Hogg, Lina Song, Edy Lim, Eugene Jeong, Tairan He, Haoru Xue, Wenli Xiao, Simon Yuen, Jan Kautz, Yan Chang, Umar Iqbal, Linxi~“Jim” Fan, and Yuke Zhu.
\newblock {SONIC}: Supersizing motion tracking for natural humanoid whole-body control.
\newblock \emph{Science Robotics}, 11\penalty0 (117), 2026.

\bibitem[Zeng et~al.(2026)Zeng, Yin, Niu, Lu, Zhong, Chen, Jia, Chen, Wang, Xu, Zhou, Li, Zhang, Wang, Yi, Lin, Pang, and Wang]{Zeng2026ScaleBFM}
Weishuai Zeng, Kangning Yin, Xiaojie Niu, Shunlin Lu, Weixiang Zhong, Jiahe Chen, Feiyu Jia, Xiao Chen, Zirui Wang, Furui Xu, Ming Zhou, Kailin Li, Weinan Zhang, He~Wang, Li~Yi, Dahua Lin, Jiangmiao Pang, and Jingbo Wang.
\newblock Scaling behavior foundation model for humanoid robots.
\newblock \emph{arXiv preprint arXiv:2607.15163}, 2026.

\bibitem[Li et~al.(2026)Li, Luo, Zhang, Dai, Kanervisto, Tirinzoni, Weng, Kitani, Guzek, Touati, Lazaric, Pirotta, and Shi]{Li2026BFMZero}
Yitang Li, Zhengyi Luo, Tonghe Zhang, Cunxi Dai, Anssi Kanervisto, Andrea Tirinzoni, Haoyang Weng, Kris Kitani, Mateusz Guzek, Ahmed Touati, Alessandro Lazaric, Matteo Pirotta, and Guanya Shi.
\newblock {BFM-Zero}: A promptable behavioral foundation model for humanoid control using unsupervised reinforcement learning.
\newblock In \emph{Proc. of the International Conf. on Learning Representations (ICLR)}, 2026.

\bibitem[Jiang et~al.(2026)Jiang, Zhong, Zhang, Liu, Pan, Yang, Wang, Li, and Chen]{Jiang2026PingpongDuo}
Haoran Jiang, Yichao Zhong, Kai Zhang, Cheng Liu, Yixuan Pan, Zhenjie Yang, Jingbo Wang, Tianyu Li, and Li~Chen.
\newblock {PingpongDuo}: Agile humanoid table tennis doubles with safe team coordination.
\newblock \emph{arXiv preprint arXiv:2610.20579}, 2026.

\bibitem[Ren et~al.(2026)Ren, Li, Zhang, Fu, Jiang, Pan, Zeng, Huang, Guo, Lu, Li, Wang, Chen, Li, and Luo]{Ren2026SMASH}
Junli Ren, Yinghui Li, Kai Zhang, Penglin Fu, Haoran Jiang, Yixuan Pan, Guangjun Zeng, Tao Huang, Weizhong Guo, Peng Lu, Tianyu Li, Jingbo Wang, Li~Chen, Hongyang Li, and Ping Luo.
\newblock {SMASH}: Mastering scalable whole-body skills for humanoid ping-pong with egocentric vision.
\newblock \emph{IEEE Transactions on Robotics (TRO)}, 2026.

\bibitem[Zhong et~al.(2026)Zhong, Lu, Lu, Tang, Mai, Pan, Li, Chen, Wang, Li, Lu, and Li]{Zhong2026RoboNaldo}
Yichao Zhong, Yidan Lu, Yuhang Lu, Tianyang Tang, Haoguang Mai, Yixuan Pan, Tianyu Li, Li~Chen, Jingbo Wang, Zhongyu Li, Peng Lu, and Hongyang Li.
\newblock {RoboNaldo}: Accurate, stable and powerful humanoid soccer shooting via motion-guided curriculum reinforcement learning.
\newblock In \emph{Proc. Conf. on Robot Learning (CoRL)}, 2026.

\bibitem[Shi et~al.(2026)Shi, Peng, Chen, Jiang, Li, Huang, Luo, Li, and Chen]{shi2026egohumanoid}
Modi Shi, Shijia Peng, Jin Chen, Haoran Jiang, Tianyu Li, Di~Huang, Ping Luo, Hongyang Li, and Li~Chen.
\newblock {EgoHumanoid}: Unlocking in-the-wild loco-manipulation with robot-free egocentric demonstration.
\newblock In \emph{Proc. Robotics: Science and Systems (RSS)}, 2026.

\bibitem[Chen et~al.(2026)Chen, Jiang, Xu, Shi, Peng, Chen, Li, Xu, Chen, Hoi, and Li]{Chen2026EgoHumanoidV2}
Jin Chen, Yiming Jiang, Chongyang Xu, Modi Shi, Shijia Peng, Li~Chen, Tianyu Li, Mu~Xu, Yilun Chen, Steven Hoi, and Hongyang Li.
\newblock {EgoHumanoid-V2}: Human-to-humanoid transfer of coordinated whole-body skills for loco-manipulation.
\newblock \emph{arXiv preprint arXiv:2609.37181}, 2026.

\bibitem[Araujo et~al.(2026)Araujo, Ze, Xu, Wu, and Liu]{Araujo2026GMR}
Joao~Pedro Araujo, Yanjie Ze, Pei Xu, Jiajun Wu, and C.~Karen Liu.
\newblock Retargeting matters: {General Motion Retargeting} for humanoid motion tracking.
\newblock In \emph{Proc. IEEE International Conf. on Robotics and Automation (ICRA)}, 2026.

\bibitem[Cao et~al.(2026)Cao, Fang, Kwon, Yu, Ma, Tan, Zhou, Du, Gu, Gao, Dai, Han, and Xu]{Cao2026UMR}
Hanyang Cao, Yuetong Fang, Taesoo Kwon, Runyi Yu, Ji~Ma, Jing Tan, Yangchen Zhou, Baoze Du, Yi~Gu, Yukang Gao, Ruoli Dai, Lei Han, and Renjing Xu.
\newblock {Unified Motion Retargeting} for humanoids with learned point cloud correspondence.
\newblock \emph{arXiv preprint arXiv:2609.02134}, 2026.

\bibitem[Loper et~al.(2015)Loper, Mahmood, Romero, Pons-Moll, and Black]{SMPL:2015}
Matthew Loper, Naureen Mahmood, Javier Romero, Gerard Pons-Moll, and Michael~J. Black.
\newblock {SMPL}: A skinned multi-person linear model.
\newblock \emph{ACM Trans. on Graphics}, 34\penalty0 (6), 2015.

\bibitem[Pan et~al.(2026)Pan, Qiao, Chen, Chitta, Pan, Mai, Bu, Zhao, Zheng, Luo, and Li]{Pan2025AMS}
Yixuan Pan, Ruoyi Qiao, Li~Chen, Kashyap Chitta, Liang Pan, Haoguang Mai, Qingwen Bu, Hao Zhao, Cunyuan Zheng, Ping Luo, and Hongyang Li.
\newblock {Agility Meets Stability}: Versatile humanoid control with heterogeneous data.
\newblock In \emph{Proc. IEEE International Conf. on Robotics and Automation (ICRA)}, 2026.

\bibitem[Sferrazza et~al.(2024)Sferrazza, Huang, Lin, Lee, and Abbeel]{Sferrazza2024HumanoidBench}
Carmelo Sferrazza, Dun-Ming Huang, Xingyu Lin, Youngwoon Lee, and Pieter Abbeel.
\newblock {HumanoidBench}: Simulated humanoid benchmark for whole-body locomotion and manipulation.
\newblock \emph{arXiv preprint arXiv:2403.10506}, 2024.

\bibitem[Wang et~al.(2026)Wang, Xie, Yang, Wang, Yuan, Fang, Feng, Wang, Chen, Chen, Wu, Xu, Chen, Li, Zeng, and Xu]{Wang2026HumanoidArena}
Taowen Wang, Zikang Xie, Bin Yang, Yunheng Wang, Zizhao Yuan, Yuetong Fang, Yixiao Feng, Yichi Wang, Xingyu Chen, Haodong Chen, Qiwei Wu, Weisheng Xu, Lihan Chen, Lusong Li, Zecui Zeng, and Renjing Xu.
\newblock {HumanoidArena}: Benchmarking egocentric hierarchical whole-body learning.
\newblock \emph{arXiv preprint arXiv:2606.17833}, 2026.

\bibitem[{MotionChallenge Team}(2026)]{motionchallenge2026}
{MotionChallenge Team}.
\newblock {MotionChallenge}: An open, reproducible benchmark for humanoid motion tracking.
\newblock \url{https://motionchallenge-project.github.io/motionchallenge/}, 2026.

\bibitem[Liu et~al.(2026)Liu, Qi, Zeng, Yu, Guan, Zhang, Chen, Liang, Li, Lin, Yu, Zhang, Wang, and Yi]{liu2026humantrackercomprehensivehumanalignedmotion}
Dairu Liu, Zekun Qi, Jiayu Zeng, Ruixi Yu, Yu~Guan, Yintianrun Zhang, Xuchuan Chen, Sikai Liang, Zekai Li, Chenghuai Lin, Xinqiang Yu, Wenyao Zhang, He~Wang, and Li~Yi.
\newblock {HumanTracker}: Towards comprehensive and human-aligned motion tracking benchmark.
\newblock \emph{arXiv preprint arXiv:2608.13555}, 2026.

\bibitem[Tessler et~al.(2024)Tessler, Guo, Nabati, Chechik, and Peng]{tessler2024maskedmimic}
Chen Tessler, Yunrong Guo, Ofir Nabati, Gal Chechik, and Xue~Bin Peng.
\newblock {MaskedMimic}: Unified physics-based character control through masked motion inpainting.
\newblock \emph{ACM Trans. on Graphics}, 2024.

\bibitem[He et~al.(2025)He, Xiao, Lin, Luo, Xu, Jiang, Kautz, Liu, Shi, Wang, Fan, and Zhu]{he2025hover}
Tairan He, Wenli Xiao, Toru Lin, Zhengyi Luo, Zhenjia Xu, Zhenyu Jiang, Jan Kautz, Changliu Liu, Guanya Shi, Xiaolong Wang, Linxi~Jim Fan, and Yuke Zhu.
\newblock {HOVER}: Versatile neural whole-body controller for humanoid robots.
\newblock In \emph{Proc. IEEE International Conf. on Robotics and Automation (ICRA)}, 2025.

\bibitem[{Bones Studio}(2026)]{bones_seed_dataset}
{Bones Studio}.
\newblock {BONES-SEED}: Skeletal everyday embodiment dataset.
\newblock \url{https://huggingface.co/datasets/bones-studio/seed}, 2026.
\newblock Dataset. Accessed: 2026-10-25.

\bibitem[Ara{\'u}jo et~al.(2023)Ara{\'u}jo, Li, Vetrivel, Agarwal, Wu, Gopinath, Clegg, and Liu]{araujo2023circle}
Joao~Pedro Ara{\'u}jo, Jiaman Li, Karthik Vetrivel, Rishi Agarwal, Jiajun Wu, Deepak Gopinath, Alexander~William Clegg, and Karen Liu.
\newblock {CIRCLE}: Capture in rich contextual environments.
\newblock In \emph{Proc. IEEE Conf. on Computer Vision and Pattern Recognition (CVPR)}, pages 21211--21221, 2023.

\bibitem[Li et~al.(2023)Li, Wu, and Liu]{li2023object}
Jiaman Li, Jiajun Wu, and C~Karen Liu.
\newblock Object motion guided human motion synthesis.
\newblock \emph{ACM Trans. Graph.}, 42\penalty0 (6), 2023.

\bibitem[Mahmood et~al.(2019)Mahmood, Ghorbani, Troje, Pons-Moll, and Black]{mahmood2019amass}
Naureen Mahmood, Nima Ghorbani, Nikolaus~F Troje, Gerard Pons-Moll, and Michael~J Black.
\newblock {AMASS}: Archive of motion capture as surface shapes.
\newblock In \emph{Proc. of the IEEE International Conf. on Computer Vision (ICCV)}, 2019.

\bibitem[Rempe et~al.(2026)Rempe, Petrovich, Yuan, Zhang, Peng, Jiang, Wang, Iqbal, Minor, de~Ruyter, Li, Tessler, Lim, Jeong, Wu, Hassani, Huang, Yu, Chung, Song, Dionne, Kautz, Yuen, and Fidler]{rempe2026kimodo}
Davis Rempe, Mathis Petrovich, Ye~Yuan, Haotian Zhang, Xue~Bin Peng, Yifeng Jiang, Tingwu Wang, Umar Iqbal, David Minor, Michael de~Ruyter, Jiefeng Li, Chen Tessler, Edy Lim, Eugene Jeong, Sam Wu, Ehsan Hassani, Michael Huang, Jin-Bey Yu, Chaeyeon Chung, Lina Song, Olivier Dionne, Jan Kautz, Simon Yuen, and Sanja Fidler.
\newblock Kimodo: Scaling controllable human motion generation.
\newblock \emph{arXiv preprint arXiv:2603.15546}, 2026.

\bibitem[{RoboParty Lab Team}(2026)]{RoboParty2026MimicLite}
{RoboParty Lab Team}.
\newblock {MimicLite}: Efficient and effective general humanoid motion tracking.
\newblock \url{https://github.com/EGalahad/mimic-lite}, 2026.
\newblock Technical report: \url{https://github.com/Roboparty/MimicLite/blob/main/mimic-lite.pdf}.

\bibitem[NVIDIA et~al.(2025)NVIDIA, :, Bjorck, Castañeda, Cherniadev, Da, Ding, Fan, Fang, Fox, Hu, Huang, Jang, Jiang, Kautz, Kundalia, Lao, Li, Lin, Lin, Liu, Llontop, Magne, Mandlekar, Narayan, Nasiriany, Reed, Tan, Wang, Wang, Wang, Wang, Xiang, Xie, Xu, Xu, Ye, Yu, Zhang, Zhang, Zhao, Zheng, and Zhu]{nvidia2025gr00tn1openfoundation}
NVIDIA, :, Johan Bjorck, Fernando Castañeda, Nikita Cherniadev, Xingye Da, Runyu Ding, Linxi~"Jim" Fan, Yu~Fang, Dieter Fox, Fengyuan Hu, Spencer Huang, Joel Jang, Zhenyu Jiang, Jan Kautz, Kaushil Kundalia, Lawrence Lao, Zhiqi Li, Zongyu Lin, Kevin Lin, Guilin Liu, Edith Llontop, Loic Magne, Ajay Mandlekar, Avnish Narayan, Soroush Nasiriany, Scott Reed, You~Liang Tan, Guanzhi Wang, Zu~Wang, Jing Wang, Qi~Wang, Jiannan Xiang, Yuqi Xie, Yinzhen Xu, Zhenjia Xu, Seonghyeon Ye, Zhiding Yu, Ao~Zhang, Hao Zhang, Yizhou Zhao, Ruijie Zheng, and Yuke Zhu.
\newblock {GR00T N1}: An open foundation model for generalist humanoid robots.
\newblock \emph{arXiv preprint arXiv:2503.14734}, 2025.

\bibitem[Harvey et~al.(2020)Harvey, Yurick, et~al.]{harvey2020robust}
Félix~G. Harvey, Mike Yurick, et~al.
\newblock Robust motion in-betweening.
\newblock \emph{ACM Trans. on Graphics}, 2020.

\bibitem[{NVIDIA}()]{nvidia_soma_retargeter}
{NVIDIA}.
\newblock {SOMA Retargeter}.
\newblock URL \url{https://github.com/NVIDIA/soma-retargeter}.
\newblock SOMA BVH to humanoid robot motion retargeting library.

\bibitem[Schulman et~al.(2017)Schulman, Wolski, Dhariwal, Radford, and Klimov]{schulman2017proximal}
John Schulman, Filip Wolski, Prafulla Dhariwal, Alec Radford, and Oleg Klimov.
\newblock Proximal policy optimization algorithms.
\newblock \emph{arXiv preprint arXiv:1707.06347}, 2017.

\bibitem[van~der Maaten and Hinton(2008)]{VanDerMaaten2008TSNE}
Laurens van~der Maaten and Geoffrey Hinton.
\newblock Visualizing data using {t-SNE}.
\newblock \emph{Journal of Machine Learning Research}, 9\penalty0 (86):\penalty0 2579--2605, 2008.

\bibitem[Yang et~al.(2025)Yang, Wang, Yang, Slabaugh, and Yuan]{Yang2025STaR}
Xiaohang Yang, Qing Wang, Jiahao Yang, Gregory Slabaugh, and Shanxin Yuan.
\newblock {STaR}: Seamless spatial-temporal aware motion retargeting with penetration and consistency constraints.
\newblock In \emph{Proc. of the IEEE International Conf. on Computer Vision (ICCV)}, pages 12947--12955, 2025.

\end{thebibliography}

\clearpage
\bookmarksetup{startatroot}
\phantomsection
\pdfbookmark[1]{Supplementary Materials}{supplementary-materials}
{\noindent\Large\itshape Supplementary Materials\par}

\setcounter{section}{0}
\setcounter{subsection}{0}
\setcounter{figure}{0}
\setcounter{table}{0}

\renewcommand{\thesection}{S\arabic{section}}
\renewcommand{\thesubsection}{\thesection.\arabic{subsection}}
\renewcommand{\thefigure}{S\arabic{figure}}
\renewcommand{\thetable}{S\arabic{table}}

\renewcommand{\theHsection}{supp.\arabic{section}}
\renewcommand{\theHsubsection}{\theHsection.\arabic{subsection}}
\renewcommand{\theHfigure}{suppfig.\arabic{figure}}
\renewcommand{\theHtable}{supptab.\arabic{table}}

\begingroup
\renewcommand{\topfraction}{0.9}
\renewcommand{\bottomfraction}{0.8}
\renewcommand{\textfraction}{0.1}
\section{Benchmark coverage and composition}
\label{supp:sec2:coverage}
\label{supp:sec1:subsec:analysis}
We define eligible In-house track tasks using upper-body and lower-body action descriptions to specify manipulation-related coordination demands.
Sit--stand tasks are excluded because the benchmark lacks a unified quantitative evaluation protocol for seated configurations and transitions between support contacts.
Their tags are retained in the reference set to make this scope limitation explicit in the coverage analysis.

For each eligible task, we select up to five motion-capture sequences that pass the same automated screening and human review procedures used for the in-house training data (\cref{sec4:subsec:training-data:subsubsec:qc}).
We retain all qualifying sequences when fewer than five are available and record tasks with none as unrepresented.
We assess the resulting track using the released action tags and compatible pairs of first-level and second-level tags.

\subsection{Tag and combination coverage}
\label{supp:sec2:subsec:tag-coverage}
Tag coverage is the fraction of reference tags represented in an evaluation set.
Pair coverage is the fraction of compatible pairs of first-level and second-level tags represented within the upper-body or lower-body description.
Let $\mathcal{T}$ denote the reference tag set and $\mathcal{P}_b$ the valid pair set for body description $b\in\{U,L\}$, where $U$ and $L$ denote upper and lower body.
For an evaluation set $\mathcal{D}$, let $\mathcal{T}(\mathcal{D})$ and $\mathcal{P}_b(\mathcal{D})$ denote the distinct tags and jointly represented pairs of first-level and second-level tags identified in its annotations.
Coverage is defined as
\begin{equation}
\begin{aligned}
C_{\mathrm{tag}}(\mathcal{D})
&= \frac{|\mathcal{T}(\mathcal{D})\cap\mathcal{T}|}{|\mathcal{T}|}, \\
C_{\mathrm{pair}}^{b}(\mathcal{D})
&= \frac{|\mathcal{P}_b(\mathcal{D})\cap\mathcal{P}_b|}{|\mathcal{P}_b|},
\qquad b\in\{U,L\}, \\
C_{\mathrm{pair}}(\mathcal{D})
&= \frac{\sum_{b\in\{U,L\}}|\mathcal{P}_b(\mathcal{D})\cap\mathcal{P}_b|}
{\sum_{b\in\{U,L\}}|\mathcal{P}_b|}.
\end{aligned}
\label{supp:eq:coverage}
\end{equation}
Here, $|\mathcal{T}|=43$, $|\mathcal{P}_U|=33$, and $|\mathcal{P}_L|=20$.
Overall pair coverage pools the covered counts across the 53 valid pairs, weighting each pair equally.
Coverage values are reported as percentages.
These measures characterize coverage relative to the predefined tag and pair sets.

The In-house track covers 39 of the 43 reference tags (90.70\%; \Cref{supp:tab:benchmark-coverage}).
The four uncovered tags concern torso coordination during sit--stand transitions (\texttt{U1-8}), sit--stand transitions (\texttt{L1-6}), floor sit--stand transitions (\texttt{L2-18}), and chair sit--stand transitions (\texttt{L2-19}).
It also covers 30 of 33 upper-body pairs (90.91\%) and 18 of 20 lower-body pairs (90.00\%), yielding an overall pair coverage of 48 of 53 (90.57\%).

For public sequences, GPT-6 Astra infers candidate action tags from sequence filenames using the reference tag definitions.
Public coverage estimates are computed from these filename-based candidate annotations and may be affected by incomplete or ambiguous filename descriptions.

\begin{table}[t]
\centering
\caption{\textbf{Action-tag and combination coverage across tracks and public data sources.}
Entries report covered/reference counts and percentages.
Tag and pair denominators are fixed across all evaluated sets.
Upper-body and lower-body pairs link first-level and second-level tags within the respective body description; all pairs combine these two sets.
Public (combined) reports the union of distinct tags and compatible pairs across the four public sources.}
\label{supp:tab:benchmark-coverage}
\label{supp:tab:coverage}
\footnotesize
\setlength{\tabcolsep}{4pt}
\renewcommand{\arraystretch}{1.15}
\begin{tabular}{@{}lcccc@{}}
\toprule
Track/source & Tags & \shortstack{Upper-body\\pairs} & \shortstack{Lower-body\\pairs} & All pairs \\
\midrule
\textbf{In-house} & \textbf{\shortstack{39/43\\(90.70\%)}} & \textbf{\shortstack{30/33\\(90.91\%)}} & \textbf{\shortstack{18/20\\(90.00\%)}} & \textbf{\shortstack{48/53\\(90.57\%)}} \\
BONES-SEED & \shortstack{28/43\\(65.12\%)} & \shortstack{9/33\\(27.27\%)} & \shortstack{11/20\\(55.00\%)} & \shortstack{20/53\\(37.74\%)} \\
CIRCLE & \shortstack{22/43\\(51.16\%)} & \shortstack{5/33\\(15.15\%)} & \shortstack{10/20\\(50.00\%)} & \shortstack{15/53\\(28.30\%)} \\
OMOMO & \shortstack{25/43\\(58.14\%)} & \shortstack{6/33\\(18.18\%)} & \shortstack{11/20\\(55.00\%)} & \shortstack{17/53\\(32.08\%)} \\
AMASS & \shortstack{26/43\\(60.47\%)} & \shortstack{4/33\\(12.12\%)} & \shortstack{11/20\\(55.00\%)} & \shortstack{15/53\\(28.30\%)} \\
\textbf{Public (combined)} & \textbf{\shortstack{29/43\\(67.44\%)}} & \textbf{\shortstack{11/33\\(33.33\%)}} & \textbf{\shortstack{11/20\\(55.00\%)}} & \textbf{\shortstack{22/53\\(41.51\%)}} \\
\bottomrule
\end{tabular}
\end{table}

\FloatBarrier
We use t-distributed stochastic neighbor embedding (t-SNE) ~\cite{VanDerMaaten2008TSNE} to visualize semantic and pose features from the three tracks (\cref{fig:bench-overview}B).
The projections show overlapping and distinct regions in the two feature representations and provide a qualitative comparison.

\subsection{Action-tag distributions}
\label{supp:sec2:subsec:distributions}
We compare sequence-level prevalence across the 10 upper-body and 19 lower-body second-level tags to characterize the composition of the In-house and Public tracks (\cref{supp:fig:tag-prevalence}).
The finalized In-house track contains 15,342 motion sequences across 3,247 tasks.
The In-house track emphasizes unilateral and bimanual reaching, with lower-body actions distributed more broadly across stance and movement categories than in the Public track.
By comparison, the filename-based candidate annotations for sequences in the Public track show greater concentration in arms-at-side behavior, side-by-side stance, and turning.

\begin{figure}[t]
\centering
\includegraphics[width=\linewidth]{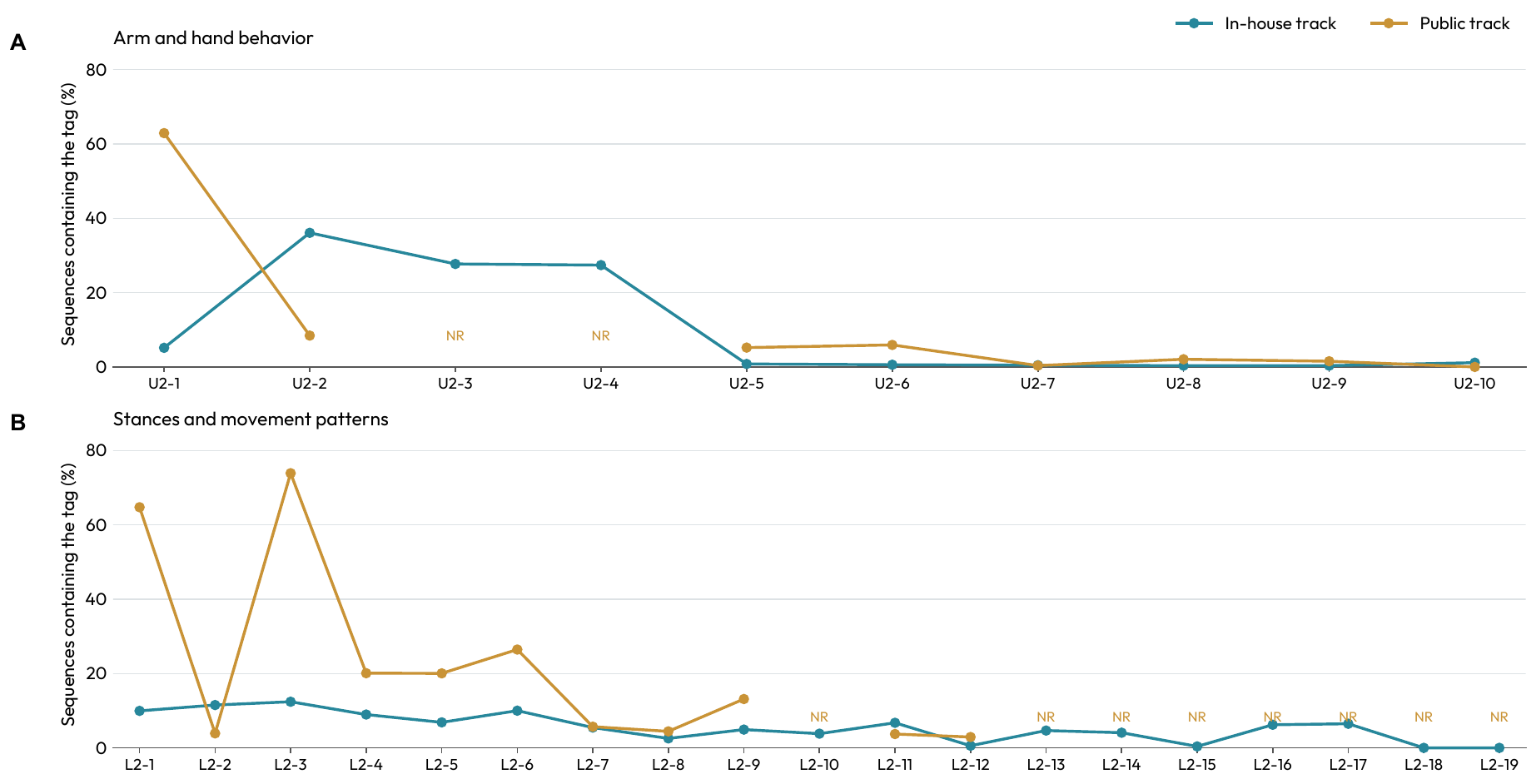}
\caption{\textbf{Prevalence of second-level action tags.}
\textbf{(A)} Arm and hand behavior.
\textbf{(B)} Stances and movement patterns.
Tag codes follow \Cref{tab:inhouse-action-tags}.
Prevalence is the number of sequences assigned a tag divided by the total number of sequences in the respective track: 15,342 for the In-house track and 2,230 for the Public track.
Each full sequence is counted once per tag and may carry multiple tags, so percentages need not sum to 100\%.
Estimates for the Public track use filename-based candidate annotations; NR denotes unresolved tags in this track, which are not plotted as zero.
Lines connect discrete categories as visual guides.}
\label{supp:fig:tag-prevalence}
\label{supp:fig:tag-dist}
\end{figure}

\subsection{Spatial coverage of benchmark tracks}
\label{supp:sec2:subsec:spatial-coverage}

Under the coordinate transformations specified below, the in-house references show broader wrist-position coverage, particularly in the stationary subset (\cref{fig:bench-overview}C).
The public references span a larger region of pelvis displacement after initial alignment.
These observations describe empirical occupancy of reference motions.

Positions are computed by forward kinematics of the G1 robot model, with X forward, Y left, and Z upward.
For upper-body workspace, we compute both wrist-yaw link origins with pelvis position and orientation fixed, retaining waist and arm motion.
For lower-body workspace, we compute both ankle-roll link origins with pelvis position and orientation fixed.
For stationary manipulation, we remove instantaneous pelvis translation and yaw while retaining roll and pitch before computing wrist-link origins.
For locomotion coverage, we align pelvis trajectories for moving motions by their initial horizontal position and heading, preserving subsequent displacement and absolute height.
The upper-body, lower-body, and stationary-manipulation views include a constant vertical offset of $0.78\,\mathrm{m}$ for visualization only.

Within each spatial comparison, the combined coordinate bounds of both tracks define a shared $64\times64\times64$ grid of rectangular voxels.
For the sequence set $\mathcal{D}$ used in a spatial comparison, sequence $i$ contains $T_i$ frames at sampling rate $f_i$ and contributes $m$ positions per frame: both limb endpoints for upper-body workspace, lower-body workspace, and stationary manipulation ($m=2$), or the pelvis for locomotion coverage ($m=1$).
The time-weighted count and normalized probability mass of voxel $v$ are
\begin{equation}
\begin{aligned}
H_{\mathcal{D}}(v)
&=\sum_{i\in\mathcal{D}}\sum_{t=1}^{T_i}\sum_{e=1}^{m}
\frac{1}{f_i}\,\mathbf{1}\!\left[\mathbf{x}_{i,t,e}\in v\right],\\
P_{\mathcal{D}}(v)
&=\frac{H_{\mathcal{D}}(v)}{\sum_u H_{\mathcal{D}}(u)}
=\frac{H_{\mathcal{D}}(v)}{m\sum_{i\in\mathcal{D}}T_i/f_i}.
\end{aligned}
\label{supp:eq:spatial-occupancy}
\end{equation}
Here, $\mathbf{x}_{i,t,e}$ is the transformed position and $\mathbf{1}[\cdot]$ is the indicator function.
All frames in the analyzed sequences contribute; left and right endpoints are combined with equal weight, and longer sequences contribute proportionally more time.
Each track is normalized independently, giving $\sum_v P_{\mathcal{D}}(v)=1$.
These values are probability masses; voxel sizes differ across spatial comparisons.

For each track, we display the smallest set of highest-probability voxels whose cumulative mass reaches at least $95\%$, without renormalization.
Displayed points are voxel centers rendered as soft circular marks.
Orange and yellow denote the In-house and Public tracks, respectively.
Opacity increases with log-scaled voxel probability using a shared range within each spatial comparison, bounded by the minimum and maximum nonzero voxel probabilities from both tracks, including voxels omitted from display.

Stationary and moving subsets are defined at the sequence level.
For the In-house track, only sequences whose first-level lower-body tag set is exactly $\{\texttt{L1-1}\}$ or $\{\texttt{L1-2}\}$ enter the stationary or moving subset, respectively.
Other and mixed tag sets contribute to the upper-body and lower-body workspaces but not to the stationary-manipulation or locomotion subsets.
For public sequences, let $Q_q$ denote the $q$th quantile over frames, $\mathbf{p}_{t,xy}$ the horizontal pelvis position, and $\tilde\psi_t$ the unwrapped pelvis yaw in radians.
We compute
\begin{equation}
\begin{aligned}
E_{xy}&=Q_{0.95}\!\left(\|\mathbf{p}_{t,xy}-\mathbf{p}_{1,xy}\|_2\right),\\
S_{\mathrm{yaw}}&=\frac{180}{\pi}\left[Q_{0.95}(\tilde\psi_t)-Q_{0.05}(\tilde\psi_t)\right].
\end{aligned}
\label{supp:eq:spatial-grouping}
\end{equation}
A public sequence is classified as moving if $E_{xy}\ge0.20\,\mathrm{m}$ or $S_{\mathrm{yaw}}\ge20$ degrees, and stationary otherwise.

\FloatBarrier
\endgroup
\Needspace{7\baselineskip}
\section{Reference fidelity evaluation}
\label{supp:sec5:reference-fidelity}
We compare wrist trajectories extracted from $H'$ and the input representation used by SONIC's SMPL encoder~\cite{luo2026sonic}, using the same human motion as a reference before motion encoding (\cref{fig:hprime-input-comparison}).
For the SMPL input, we first obtain the corresponding joint rotations from the human motion and then apply forward kinematics on a standard SMPL body to reconstruct the whole-body kinematics supplied to the encoder.
Inspired by the comparison of pairwise motion vectors for temporal consistency in STaR~\cite{Yang2025STaR}, we define unit-normalized pairwise displacement-vector discrepancy (unit-PDV) to quantify directional differences between the human reference and each encoder input.

Let $\rho\in\{H,H',S\}$ index the human reference $H$, $H'$, and the SMPL encoder input $S$, respectively, and let $T\in\{H',S\}$ index the two representations being compared.
We express wrist positions in common human chest axes and normalize by each representation's own fixed arm length:
\begin{equation}
    \bq^{\rho}(t)
    = \frac{\bC_{H}(t)^{\mathsf{T}}
    [\bp^{\rho}_{\mathrm{wrist}}(t)-\bp^{\rho}_{\mathrm{shoulder}}(t)]}{L_{\rho}}.
    \label{supp:eq:intent-coordinates}
\end{equation}
Here $\bC_H$ contains the human chest basis vectors, $\bp^{\rho}_{\mathrm{shoulder}}$ is the representation-specific shoulder origin, and $L_{\rho}$ is the fixed arm length.

Using matched start and end times, we compute and normalize the displacement vectors as follows:
\begin{equation}
    \bd^{\rho}_{ij}=\bq^{\rho}(t_j)-\bq^{\rho}(t_i),
    \qquad
    u(\bd)=\frac{\bd}{\max(\|\bd\|,\epsilon)}.
    \label{supp:eq:intent-displacements}
\end{equation}
Here $\epsilon>0$ stabilizes direction normalization.

The human displacement determines a weight shared by both encoder inputs, yielding the pairwise directional error $e_{ij}^{T}$:
\begin{equation}
    w^{H}_{ij}=\frac{\|\bd^{H}_{ij}\|^2}{\|\bd^{H}_{ij}\|^2+\tau^2},
    \qquad
    e^{T}_{ij}=w^{H}_{ij}
    \left\|u(\bd^{T}_{ij})-u(\bd^{H}_{ij})\right\|^2.
    \label{supp:eq:intent-pair-error}
\end{equation}
Here $\tau>0$ controls the downweighting of small human displacements.

Consider a window of $N$ samples starting at frame $s$ and a nonempty set of temporal lags $K\subseteq\{1,\ldots,N-1\}$.
For each lag $k\in K$, the within-window offset $i$ identifies a displacement pair spanning frames $s+i$ and $s+i+k$.
We define the unit-PDV score $E_s^T$ by averaging errors over all valid time-matched pairs within each lag and then averaging the lag-specific means equally:
\begin{equation}
    E^{T}_{s}=\frac{1}{|K|}\sum_{k\in K}
    \left[\frac{1}{N-k}
    \sum_{i=0}^{N-k-1}e^{T}_{s+i,s+i+k}\right].
    \label{supp:eq:intent-window-error}
\end{equation}

For visualization in \cref{fig:hprime-input-comparison}A, each normalized trajectory is centered by subtracting its mean position over the plotted interval, and the front-view axes use equal scales.
Panel B plots the unit-PDV scores $E_s^{H'}$ and $E_s^S$ at the window centers, with a dashed line indicating zero discrepancy.
The mean scores for this motion are $1.403773\times10^{-3}$ and $7.103112\times10^{-3}$, respectively.
Both overlays in panel C use the same human reference, sample times, and coordinate scales.
The example pairs $t_1\rightarrow t_n$ and $t_2\rightarrow t_3$ illustrate displacement vectors matched using identical start and end times.

\clearpage
\section{Sim-to-sim comparison}
\label{supp:sec3:isaacsim}
In \cref{supp:fig:isaacsim}, we present the Isaac Sim results, which show simulator-dependent differences in tracking and reachability performance relative to MuJoCo.
This sim-to-sim gap highlights the importance of evaluating performance in the real world or across simulation environments.

\begin{center}
    \begin{minipage}{\linewidth}
    \centering
    \includegraphics[width=\linewidth,height=0.86\textheight,keepaspectratio]{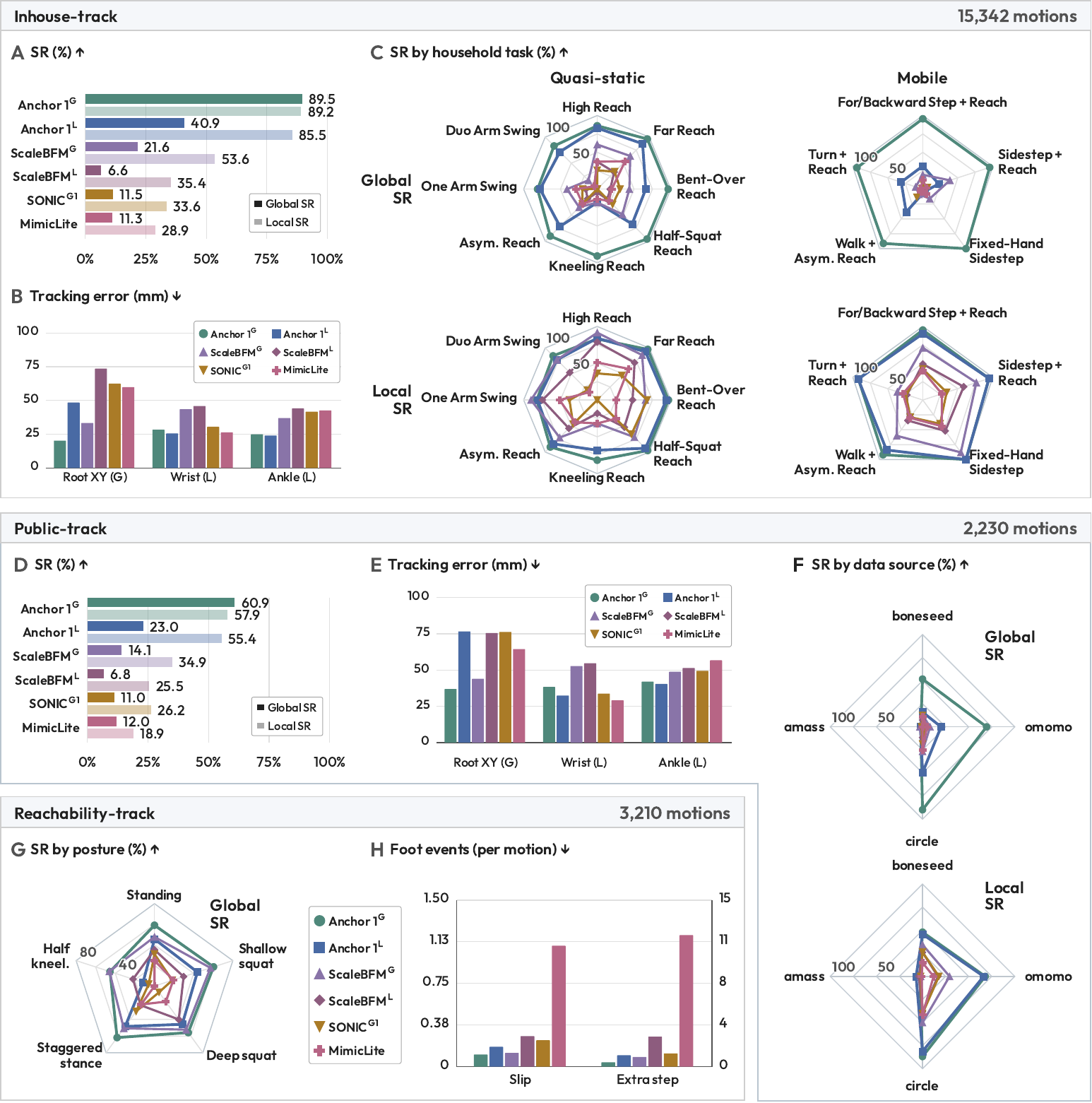}
    \captionsetup{hypcap=false}
    \captionof{figure}{\textbf{Tracking accuracy and reachability in Isaac Sim.}}
    \label{supp:fig:isaacsim}
\end{minipage}
\end{center}

\end{document}